\documentclass[10pt,twocolumn,letterpaper]{article}

\usepackage[pagenumbers]{wacv}
\usepackage{times}
\usepackage{epsfig}
\usepackage{graphicx}
\usepackage{amsmath}
\usepackage{amssymb}
\usepackage{amsthm}
\usepackage{booktabs}

\definecolor{wacvblue}{rgb}{0.21,0.49,0.74}
\usepackage[pagebackref,breaklinks,colorlinks,allcolors=wacvblue]{hyperref}

\newtheorem{rem}{Remark}

\title{Video Consistency Distance: Enhancing Temporal Consistency for \\ Image-to-Video Generation via Reward-Based Fine-Tuning}

\author{Takehiro Aoshima, Yusuke Shinohara, Byeongseon Park\\
LY Corporation\\
{\tt\small \{taaoshim, yusshino, park.byeongseon\}@lycorp.co.jp}
}

\begin{document}
\maketitle

\begin{abstract}
Reward-based fine-tuning of video diffusion models is an effective approach to improve the quality of generated videos, as it can fine-tune models without requiring real-world video datasets.
However, it can sometimes be limited to specific performances because conventional reward functions are mainly aimed at enhancing the quality across the whole generated video sequence, such as aesthetic appeal and overall consistency.
Notably, the temporal consistency of the generated video often suffers when applying previous approaches to image-to-video (I2V) generation tasks.
To address this limitation, we propose Video Consistency Distance (VCD), a novel metric designed to enhance temporal consistency, and fine-tune a model with the reward-based fine-tuning framework.
To achieve coherent temporal consistency relative to a conditioning image, VCD is defined in the frequency space of video frame features to capture frame information effectively through frequency-domain analysis.
Experimental results across multiple I2V datasets demonstrate that fine-tuning a video generation model with VCD significantly enhances temporal consistency without degrading other performance compared to the previous method.
\end{abstract}

\section{Introduction}
Video generation has witnessed significant progress over the past few years, primarily due to the rapid development of deep generative models~\cite{Goodfellow2014,grathwohl2019ffjord,ho2020denoising,Kingma2014AutoEncodingVB,song2021scorebased,NEURIPS2020_4c5bcfec,liu2022flow,NIPS2016_b1301141,pmlr-v37-sohl-dickstein15,pixelcnn2016}.
Among various approaches, diffusion-based methods have attracted particular attention owing to their ability to generate high-quality videos~\cite{opensora,chen2023seine,chen2024videocrafter2,xing2023dynamicrafter,blattmann2023stable,wang2023modelscope,wang2023lavie,bartal2024lumiere,dai2023animateanything,ho2022video}.
To further improve their specific quality, some studies proposed reward-based fine-tuning methods~\cite{prabhudesai2024videodiffusionalignmentreward,2023InstructVideo,furuta2024improvingdynamicobjectinteractions,li2024tvturbo,li2025tvturbov}.
These frameworks fine-tune a video diffusion model using a gradient-based optimization method~\cite{kingma2017adammethodstochasticoptimization,loshchilov2018decoupled}, where the gradient of the reward function is required.
Since the reward functions depend only on generated videos and conditioning data (\eg, images and texts), these methods do not require real-world video datasets for fine-tuning.
Therefore, no additional video collection, captioning, labeling, or curating is needed, and these methods are widely applicable in various scenarios.

\begin{figure}[t]
  \centering
  \footnotesize
  \tabcolsep=.3mm
  \begin{tabular}{cc}
     \rotatebox[origin=l]{90}{~Open-Sora} & \includegraphics[width=0.95\linewidth]{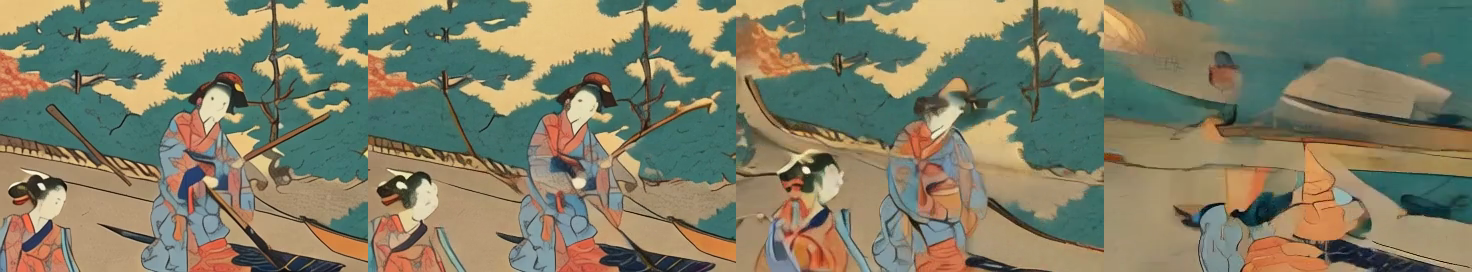} \\
     \rotatebox[origin=l]{90}{~~~~~+VCD} & \includegraphics[width=0.95\linewidth]{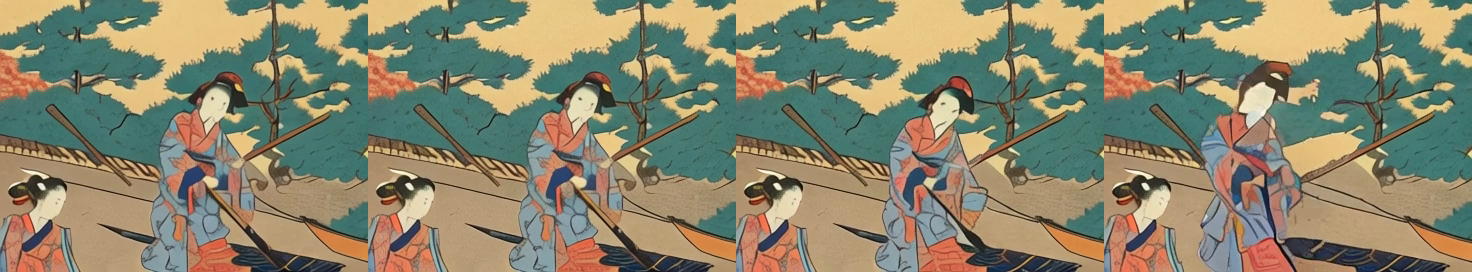} \\
     & \textit{``a person is walking''} \\
     &  \\
     \rotatebox[origin=l]{90}{~~+V-JEPA} & \includegraphics[width=0.95\linewidth]{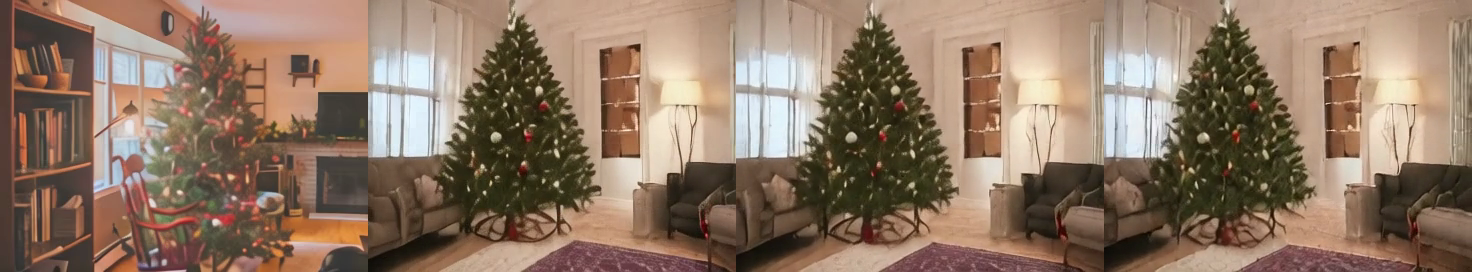} \\
     \rotatebox[origin=l]{90}{~~~~~+VCD} & \includegraphics[width=0.95\linewidth]{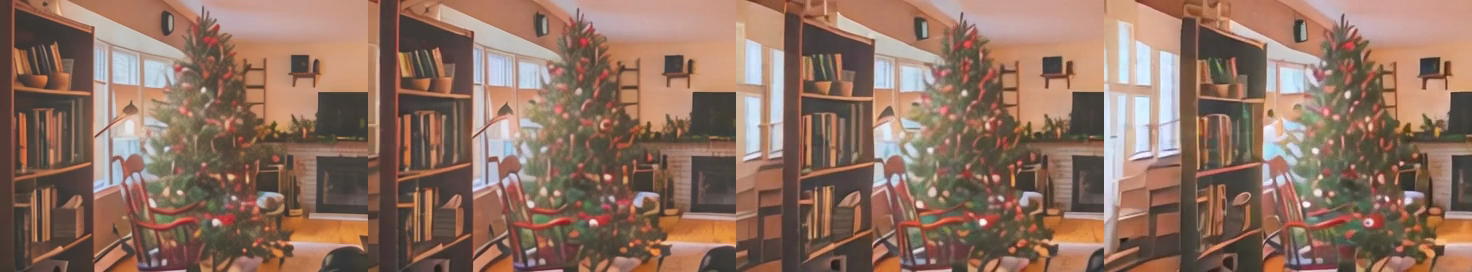} \\
      & \hspace{-5mm} \textit{``a living room with a Christmas tree and a rocking chair,} \\
      & \textit{camera pans right''} \\
      &  \\
      \rotatebox[origin=l]{90}{+Aesthetic} & \includegraphics[width=0.95\linewidth]{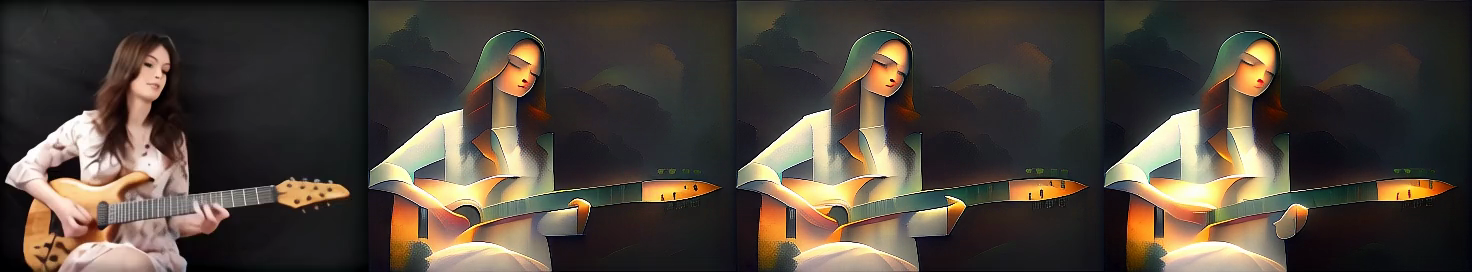} \\
      \rotatebox[origin=l]{90}{~~~~~+VCD} & \includegraphics[width=0.95\linewidth]{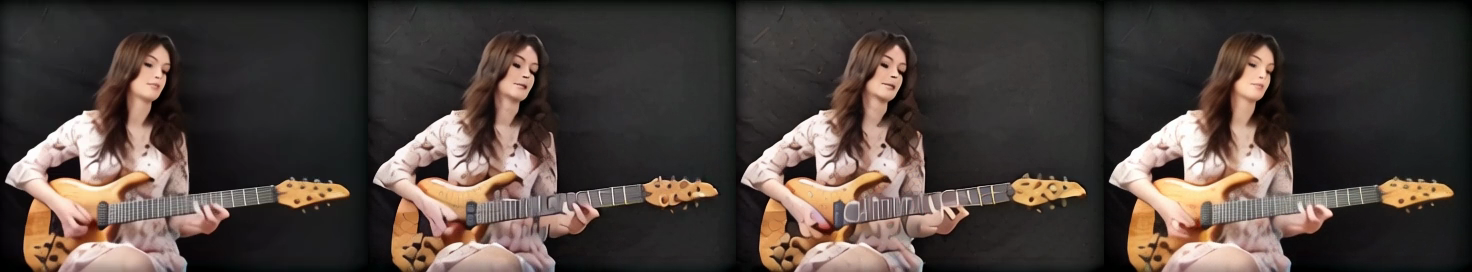} \\
      & \textit{``a woman playing guitar in front of a black screen''}
  \end{tabular}
  \caption{The examples that previous methods~\cite{opensora,prabhudesai2024videodiffusionalignmentreward} fail to preserve temporal consistency in I2V generation.
  On the top, we compare the pre-trained Open-Sora~\cite{opensora} with its generated by a model fine-tuned with VCD (+VCD).
  Open-Sora shows a significant collapse in the last frame.
  In the middle and bottom, we compare VADER, where two kinds of reward functions are employed, and our approach.
  In the middle, we present videos generated by Open-Sora models fine-tuned with two reward functions: V-JEPA~\cite{bardes2024revisiting} (+V-JEPA) and VCD.
  +V-JEPA significantly alters the various attributes of the conditioning image.
  On the bottom, we present videos generated by Open-Sora models fine-tuned with the LAION Aesthetic predictor~\cite{aesthetic} (+Aesthetic) and VCD.
  +Aeshtetic significantly alters the style of the conditioning image.
  In contrast to previous methods, ours generates temporally consistent videos relative to the conditioning images.
  }
  \label{fig:unconsistent}
\end{figure}

Although these approaches efficiently improved generated video qualities, they overlooked temporal consistency for image-to-video (I2V) generations, where preserving attributes of the conditioning image is essential (see the bottom part of Fig.~\ref{fig:unconsistent}).
Consequently, the conventional methods struggled to produce temporally consistent videos in I2V generation.
VADER~\cite{prabhudesai2024videodiffusionalignmentreward} attempted to address this limitation by employing a video feature extractor, V-JEPA~\cite{bardes2024revisiting}, as a reward function.
However, V-JEPA extracts the global features from the entire video frames without explicitly referencing the conditioning image.
Therefore, this approach struggled to cohere crucial style and object-related attributes of the conditioning image across frames, as shown in the middle part of Fig.~\ref{fig:unconsistent}.

In this paper, we propose a novel metric, namely \textit{Video Consistency Distance (VCD)}, and integrate it into a reward-based fine-tuning framework to improve temporal consistency for I2V generation.
VCD is defined as the distance between the conditioning image and a generated frame.
To enhance temporal consistency through reward-based fine-tuning, VCD should be designed to remain low when differences between the conditioning image and a generated frame are from natural motion, avoiding unnatural shifts in style or object appearance.
Conversely, it produces high values when it detects pronounced deviations due to unnatural changes, effectively identifying discrepancies that undermine temporal coherence.
To satisfy this requirement, we utilize the distance in the frequency domain of frame features, inspired by the findings of Ni \etal~\cite{ni2024misalignment} on feature frequency components in image transformation tasks.
This design helps VCD to capture frame attributes efficiently.
We validate our approach using two state-of-the-art diffusion-based video generation models, Open-Sora~\cite{opensora} and Wan2.1-1.3B-I2V~\cite{wan2025}, on three datasets: I2V-Bench~\cite{ren2024consisti2v}, VBench-I2V~\cite{huang2023vbench,huang2024vbench++}, and AI-ArtBench~\cite{silva2024artbrainexplainableendtoendtoolkit}.
Our experimental results demonstrate that the models fine-tuned with VCD generate more temporally consistent videos without degrading other qualities compared with the previous approach~\cite{prabhudesai2024videodiffusionalignmentreward}.

The contributions of this work are as follows.
\begin{enumerate}
  \item For enhancing temporal consistency in I2V generation, we introduce a novel metric, VCD, and incorporate it into a reward-based fine-tuning framework.
  Since VCD measures how naturally a generated frame moves relative to a conditioning image, it effectively improves the temporal consistency performance of an I2V generation model.
   \item We evaluate our approach on two state-of-the-art video generation models using diverse datasets.
   The experimental results show substantial improvements in temporal consistency without degrading other performance.
\end{enumerate}

\section{Related Work}
\subsection{Temporal Consistency for Video Generation Models}
As shown in the top part of Fig.~\ref{fig:unconsistent}, existing pre-trained video diffusion models sometimes fail to generate temporally consistent videos relative to the conditioning image.
Besides fine-tuning, enhancing temporal consistency in video generation has been explored through various strategies.
One notable line of research focused on preserving specific attributes, such as human face identity, by specializing in face-centric methods~\cite{yuan2024identitypreservingtexttovideogenerationfrequency,zhang2025magicmirroridpreservedvideo,anand2025ipfacediffidentitypreservingfacialvideo,ma2024magicme,zhang2025fantasyidfaceknowledgeenhanced}.
For example, Zhang \etal~\cite{zhang2025magicmirroridpreservedvideo} introduced a method for generating face identity-preserved videos by leveraging a face identity extractor~\cite{Deng_2019_CVPR}.
Although these approaches achieved impressive results in preserving specific attributes, their specialization makes them less adaptable to broader scenarios that involve diverse objects or backgrounds.

Another line of research attempted to enhance temporal consistency under specific conditions~\cite{zhang2025trainingfreemotionguidedvideogeneration,hu2023animateanyone}.
For instance, Zhang \etal~\cite{zhang2025trainingfreemotionguidedvideogeneration} proposed to generate a temporally consistent video using the motion trajectory.
However, their reliance on explicit motion cues limits applicability to I2V generation tasks, in which motion information may be absent or incomplete.

Further studies attempted to enhance temporal consistency by adding extra computation during the inference process~\cite{wu2023freeinit,xia2024unictrl,ren2024consisti2v,namekata2025sgiv}.
For example, Wu \etal~\cite{wu2023freeinit} proposed to enhance temporal consistency by iteratively refining an initial noise using the fourier transforms.
Although these techniques showed promising results, this iterative process required multiple denoising processes, increasing inference time.
Their increased inference time poses practical challenges in real-world applications.
Ren \etal~\cite{ren2024consisti2v} proposed FrameInit, which does not require a large additional inference time.
However, since its generation quality depends on the baseline model, it struggles to generate videos in unseen domains.

In this work, we aim to enhance temporal consistency by fine-tuning a model, without imposing specialized attributes/conditions or adding extra computations during the inference process.

\subsection{Fine-Tuning Diffusion Models}
Besides temporal consistency, practical applications of diffusion models often impose other specific requirements, such as text alignment and human preference.
To satisfy these requirements, previous studies proposed fine-tuning methods for diffusion models using Direct Preference Optimization (DPO)~\cite{rafailov2024directpreferenceoptimizationlanguage} or policy/reward-based frameworks~\cite{prabhudesai2024videodiffusionalignmentreward,2023InstructVideo,furuta2024improvingdynamicobjectinteractions,black2024training,clark2024directly,dong2023raft,prabhudesai2024aligning,xu2023imagereward,liu2024videodpo,wallace2023diffusionmodelalignmentusing,jiang2025huvidpoenhancingvideogenerationdirect,li2024tvturbo,li2025tvturbov}.

Some research employed DPO to fine-tune diffusion models~\cite{wallace2023diffusionmodelalignmentusing,liu2024videodpo}.
Specifically, to improve various performances, including temporal consistency, simultaneously, Liu \etal~\cite{liu2024videodpo} proposed VideoDPO, which employs a comprehensive video generation evaluation method~\cite{huang2023vbench}.
However, the temporal consistency metric employed by VideoDPO did not explicitly account for the conditioning image.
Moreover, due to the multiple metrics included, it cannot directly guarantee improved temporal consistency.

Other studies adopted reward-based frameworks~\cite{prabhudesai2024videodiffusionalignmentreward,2023InstructVideo,furuta2024improvingdynamicobjectinteractions,black2024training,clark2024directly,dong2023raft,prabhudesai2024aligning,xu2023imagereward,li2024tvturbo,li2025tvturbov}.
These approaches typically used pre-trained models, such as human preference models~\cite{wu2023human,kirstain2023pickapicopendatasetuser}, or large language models~\cite{10.5555/3600270.3602446} as a reward function to better align with practical applications.
For example, VADER~\cite{prabhudesai2024videodiffusionalignmentreward} proposed a reward-based fine-tuning framework for video diffusion models.
For improving specific qualities of generated videos, it is flexible with reward function options, such as HPS~\cite{wu2023human}, PickScore~\cite{kirstain2023pickapicopendatasetuser}, LAION Aesthetic predictor~\cite{aesthetic}, and V-JEPA~\cite{bardes2024revisiting}.
Although these reward functions effectively enhanced specific qualities, such as perception or aesthetics, employing most of them directly for I2V generation causes undesirable results (see the bottom part of Fig.~\ref{fig:unconsistent} and Fig.~\ref{fig:appendix-results-others} in Appendix~\ref{sec:appendix-other-reward-functions}).
This is because these reward functions enhance perceptual or aesthetic quality, which diverges from preserving temporal consistency relative to a conditioning image.
In the reward functions proposed by VADER, V-JEPA was employed to enhance temporal consistency by predicting a complete video from partially masked frames.
However, since it accounts for overall consistency across all generated frames, a video diffusion model fine-tuned with V-JEPA struggles to preserve temporal consistency relative to the conditioning image (see the middle part of Fig.~\ref{fig:unconsistent}).

We solve this problem by proposing a metric that calculates a distance between the conditioning image and a generated frame, and fine-tuning a model with it.

\section{Method}

\begin{figure*}[t]
\centering
\includegraphics[width=1\linewidth]{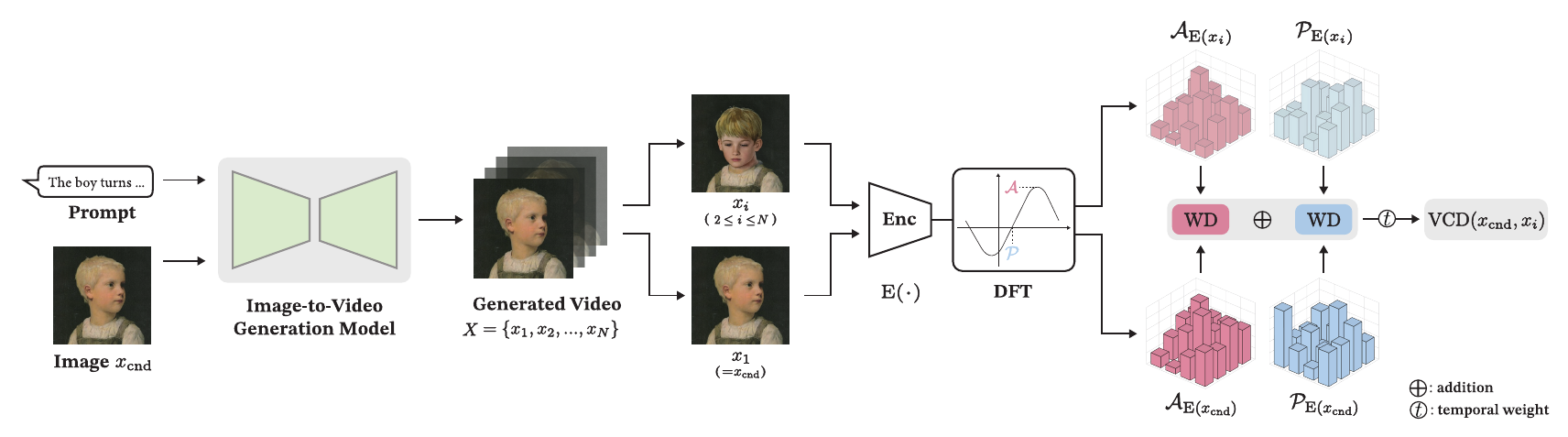}
\caption{
The overview of the proposed method.
Given the generated video $X$, we randomly sample $x_i$ to calculate Video Consistency Distance (VCD) between the conditioning image $x_{\mathrm{cnd}}$ and $x_i$.
VCD utilizes Discrete Fourier Transform (DFT)-based frequency component extraction and temporal weighting to preserve attributes of the conditioning image $x_{\mathrm{cnd}}$ while providing appropriate motion.
}
\label{fig:framework}
\vspace*{-3mm}
\end{figure*}

\subsection{Video Consistency Distance}
\label{subsec:vcd}
For a reward function to enhance temporal consistency relative to the conditioning image, it should yield a small value when the differences between the conditioning image and a generated frame are due solely to natural movement, without unnatural changes in global or local attributes.
In contrast, it should yield a large value when there are significantly unnatural changes.
For example, in the bottom part of Fig.~\ref{fig:unconsistent}, it yields a small value to +VCD since the generated video does not include unnatural movement, and yields a large value to +Aesthetic by penalizing significant style changes.
We design such a reward function inspired by the previous work~\cite{ni2024misalignment} that targets the misaligned image transformation task.

For image transformation tasks, such as image enhancement and super-resolution, Ni \etal~\cite{ni2024misalignment} proposed Frequency Distribution Loss (FDL), which computes the distribution distance between two image features in the frequency domain, defined as
\begin{equation}
\label{eq:fdl}
    \mathcal{L}_{\mathrm{FDL}} (U, V) = D(\mathcal{A}_{\mathrm{E}(U)}, \mathcal{A}_{\mathrm{E}(V)}) + \alpha D(\mathcal{P}_{\mathrm{E}(U)}, \mathcal{P}_{\mathrm{E}(V)}),
\end{equation}
where $U, V$ are images, $D$ is a distance function between two probabilistic distributions, $\mathrm{E}$ is an image encoder, $\mathcal{A}_s=|\mathcal{F}\circ s|$ and $\mathcal{P}_s=\angle(\mathcal{F}\circ s)$ are amplitude and phase of the spectrum of signal $s$, where $\mathcal{F}$ denotes the Discrete Fourier Transform (DFT), and $\alpha$ is a scaler weight, respectively.
Ni \etal demonstrated that FDL effectively handles geometric misalignments (\eg, object shifts) in training data.
The key ideas of FDL for handling misalignments are (1) calculating a distribution distance in frequency space and (2) using frequency components of an image feature.
Since the proposed metric should also be robust to geometric variation, we anticipate that these key ideas will also be effective for our goal.
We examine the influence of FDL's key ideas on enhancing temporal consistency for I2V generation models.

Research on handling geometric misalignments for image transformation frequently employed the Wasserstein Distance (WD) due to its resilience to geometric shifts~\cite{elnekave2022generating,nguyen2021distributional,zhang2019shiftinvar}.
Ni \etal showed through experimental analysis that calculating the WD in the frequency domain significantly enhances the preservation of local attributes (\eg, object shapes and edges) in transformed images.
This improvement is likely attributable to the fact that frequency components provide a more comprehensive representation of image features, thereby facilitating more accurate transformations.
Ni \etal also observed that the amplitude components of various image features capture global attributes (\eg, illumination and color), whereas the phase components capture local attributes.
In the context of I2V generation, these insights suggest that measuring the WD in the frequency domain of frame features can help to enhance temporal consistency relative to the conditioning image.
Based on the above observations, we propose the following remarks:

\begin{rem}
Leveraging the WD between a conditioning image and each generated frame in the frequency domain as a reward function is highly effective for fine-tuning I2V generation models.
This approach notably contributes to the preservation of local attributes of the conditioning image throughout the generated video sequence.
\end{rem}
\begin{rem}
Incorporating frequency components extracted from various feature representations as a reward function significantly improves the preservation of global and local attributes of the conditioning image in the generated video.
It facilitates a more coherent and visually consistent output across all generated frames.
\end{rem}

Motivated by these remarks, we define VCD that leverages the frequency components of frame features.
We show the overview of the proposed method in Fig.~\ref{fig:framework}.
Formally, given a generated video $X=\{x_1, x_2, ..., x_N\}$, we define VCD between the conditioning image $x_{\mathrm{cnd}}$ and $i$-th frame $x_i$ as
\begin{equation}
\label{eq:vcd}
\begin{split}
    \mathrm{VCD}(x_{\mathrm{cnd}}, x_i) = \frac{N-i+1}{N} ( & \mathrm{WD}(\mathcal{A}_{\mathrm{E}(x_{\mathrm{cnd}})}, \mathcal{A}_{\mathrm{E}(x_i)}) \\
    & + \mathrm{WD}(\mathcal{P}_{\mathrm{E}(x_{\mathrm{cnd}})}, \mathcal{P}_{\mathrm{E}(x_i)})),
\end{split}
\end{equation}
where $1\leq i \leq N$.
In I2V generation, a conditioning image may correspond to the $i$-th frame, and VCD can be applied in all such cases by definition.
In our experiments, a conditioning image is used as the first frame of the generated video.
Therefore, we set $x_{\mathrm{cnd}} = x_1$ and $2\leq i \leq N$.
To prevent generating a still image by over-approximating $x_{\mathrm{cnd}}$ and $x_i$, we introduce a temporal weight $\frac{N-i+1}{N}$ for the $i$-th frame.
For calculating WD, we calculate the empirical distribution by aggregating all amplitude and all phase coefficients across spatial positions and channels after applying DFT.
We employ Sliced Wasserstein Distance~\cite{Heitz_2021_CVPR} for computational efficiency.
In this work, we use shallow layers of VGG19~\cite{simonyan2015deepconvolutionalnetworkslargescale} (\textit{Relu\_1\_1}, \textit{Relu\_2\_1}, \textit{Relu\_3\_1}, \textit{Relu\_4\_1}, and \textit{Relu\_5\_1}) as an image encoder $E$ to extract various image features.
While other modern image encoders (\eg, DINOv2~\cite{Jose_2025_CVPR} and CLIP~\cite{Radford2021LearningTV}) are adaptable, we employ VGG19 for its simplicity and lightweight.
VCD becomes small if the differences between the conditioning image and a generated frame stem primarily from natural motion.
As a result, minimizing VCD encourages temporal consistency within the generated video.

\subsection{Fine-Tuning Framework}
\label{subsec:framework}
Although VCD is adaptable for any I2V generation model, this study focuses on diffusion-based models for their ability to generate a high-quality video.

Let $p_{\theta}, R, c$ represent a pre-trained I2V diffusion model with parameters $\theta$, a reward function, and conditioning data (an image or image-text pair for I2V generation task), respectively.
We can fine-tune a pre-trained I2V diffusion model $p_{\theta}$ by maximizing $J(\theta)$ where
\begin{equation}
\label{eq:fine-tune}
    J(\theta) = \mathbb{E}_{X\sim p_{\theta}(X|c)}[R(X, c)].
\end{equation}
Using the gradient of the reward function $\nabla_\theta R$, we can optimize $J(\theta)$ with a gradient-based optimization method, such as Adam~\cite{kingma2017adammethodstochasticoptimization} or AdamW~\cite{loshchilov2018decoupled}.
By optimizing $J(\theta)$ in Eq.~\ref{eq:fine-tune}, where VCD serves as the reward function $R$, we can enhance temporal consistency for an I2V generation model.
The gradient of the reward function $\nabla 
 \mathrm{VCD}(x_{\mathrm{cnd}}, x_i)$ is calculated with the conditioning image $x_{\mathrm{cnd}}$ and a generated frame $x_i$.
Therefore, a pre-trained I2V diffusion model $p_{\theta}$ can be fine-tuned without video datasets.
However, fine-tuning all parameters $\theta$ by backpropagating through every sequential denoising step consumes an enormous memory cost.
To reduce this memory consumption, techniques such as LoRA~\cite{hu2022lora} and Cache~\cite{ma2023deepcache} are employable.
Note that this fine-tuning framework is available for any video diffusion model without depending on the model architecture.

\begin{figure}[t]
\centering
\footnotesize
\tabcolsep=.5mm
\begin{tabular}{cc}
\rotatebox[origin=l]{90}{~~Open-Sora} & \includegraphics[width=0.93\linewidth]{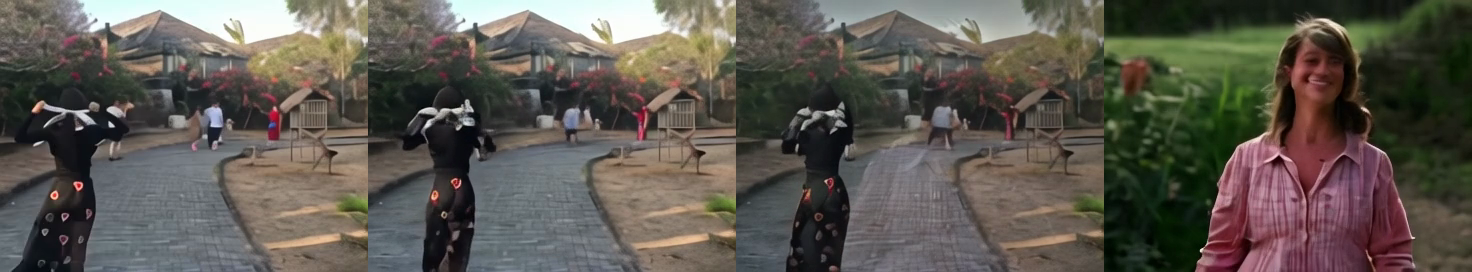} \\
 \rotatebox[origin=l]{90}{~~~+V-JEPA} & \includegraphics[width=0.93\linewidth]{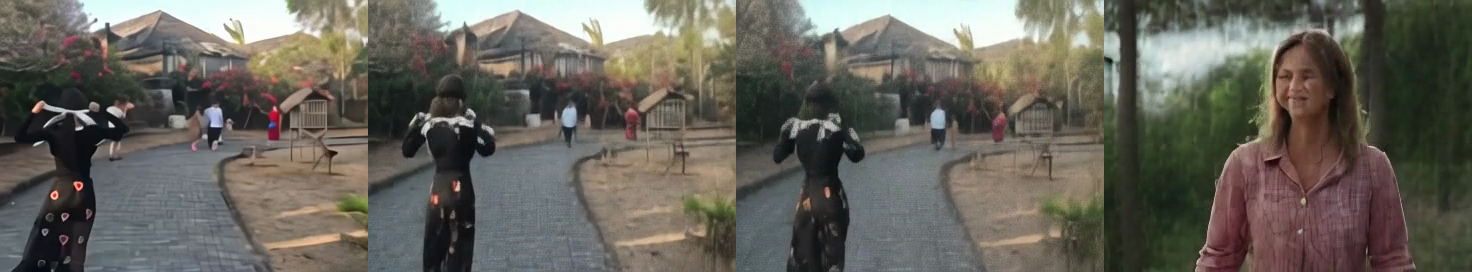} \\
 \rotatebox[origin=l]{90}{~~~+VCD} & \includegraphics[width=0.93\linewidth]{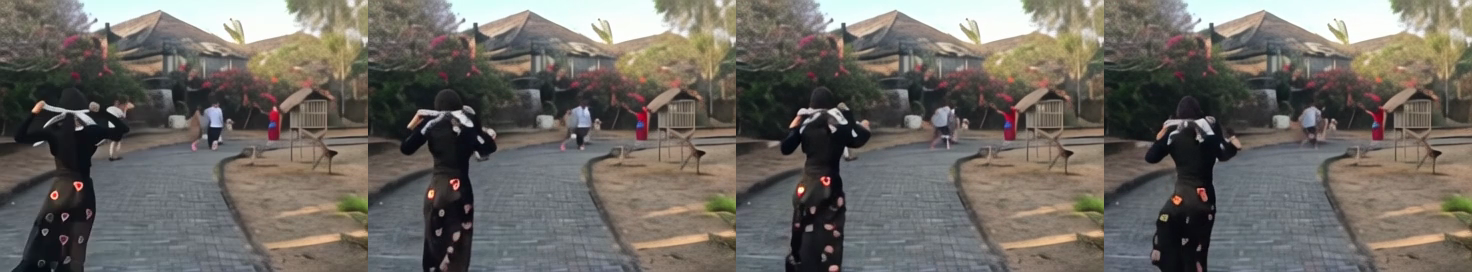} \\
 & \textit{``a woman was walking happily on the farm''} \\
  & \\
 \rotatebox[origin=l]{90}{~~~~~Wan} & \includegraphics[width=0.93\linewidth]{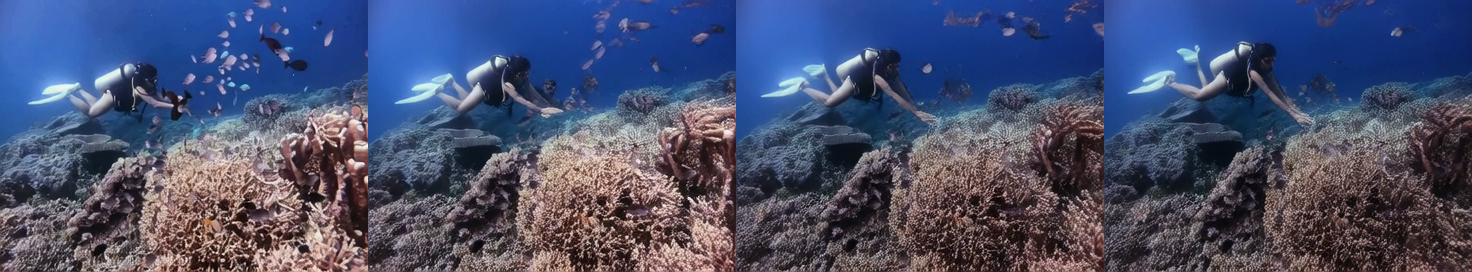} \\
 \rotatebox[origin=l]{90}{~~~+V-JEPA} & \includegraphics[width=0.93\linewidth]{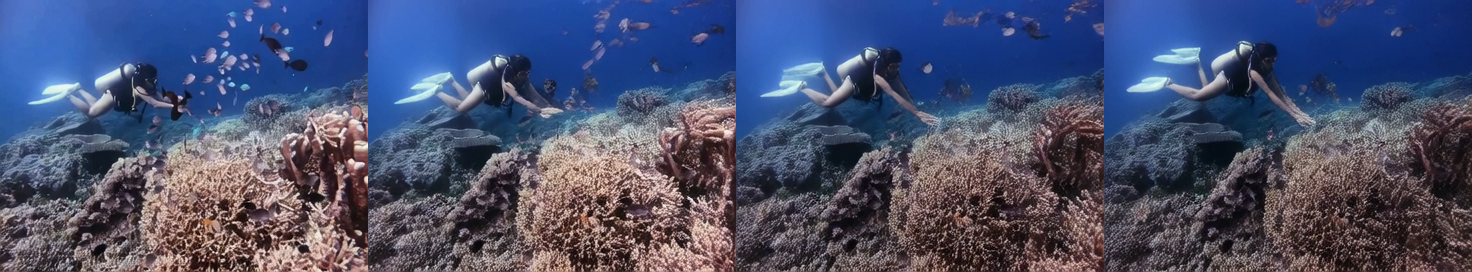} \\
 \rotatebox[origin=l]{90}{~~~+VCD} & \includegraphics[width=0.93\linewidth]{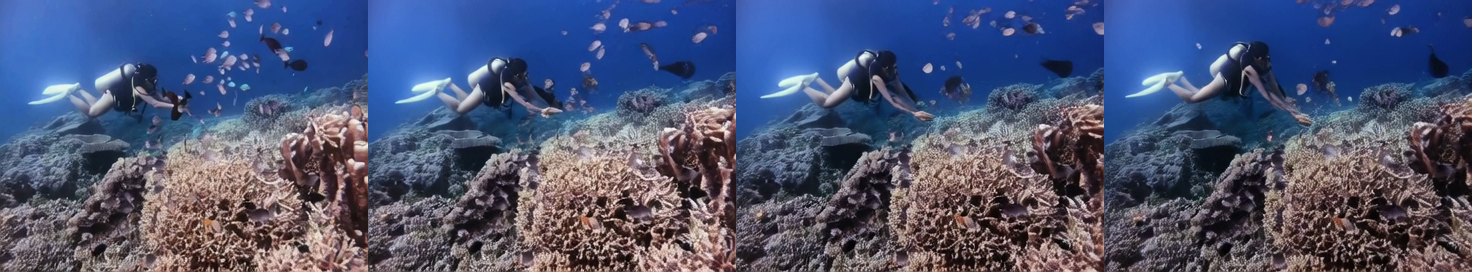} \\
 & \textit{``a woman is diving on the sea floor''}
\end{tabular}
\caption{
Results of video generation with I2V-Bench.
We provide text prompts below the figures.
}
\vspace*{-3mm}
\label{fig:results-opensora-i2v_bench}
\end{figure}

\begin{figure}[t]
\centering
\footnotesize
\tabcolsep=.5mm
\begin{tabular}{cc}
\rotatebox[origin=l]{90}{~~Open-Sora} & \includegraphics[width=0.93\linewidth]{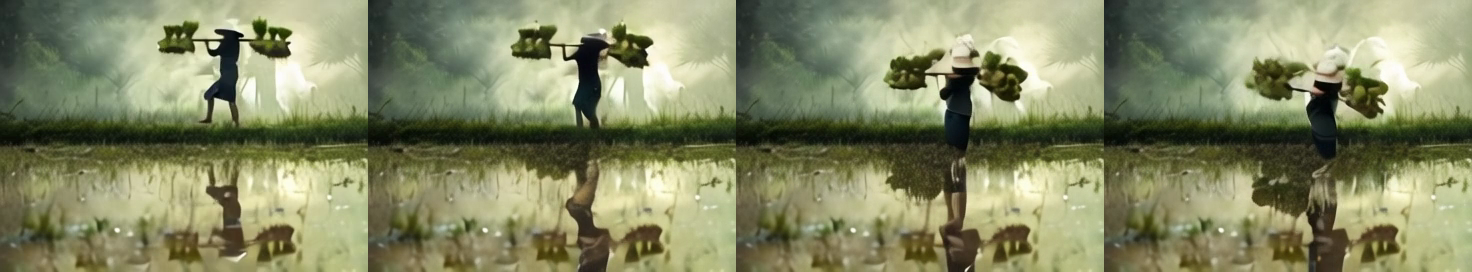} \\
\rotatebox[origin=l]{90}{~~~+V-JEPA} & \includegraphics[width=0.93\linewidth]{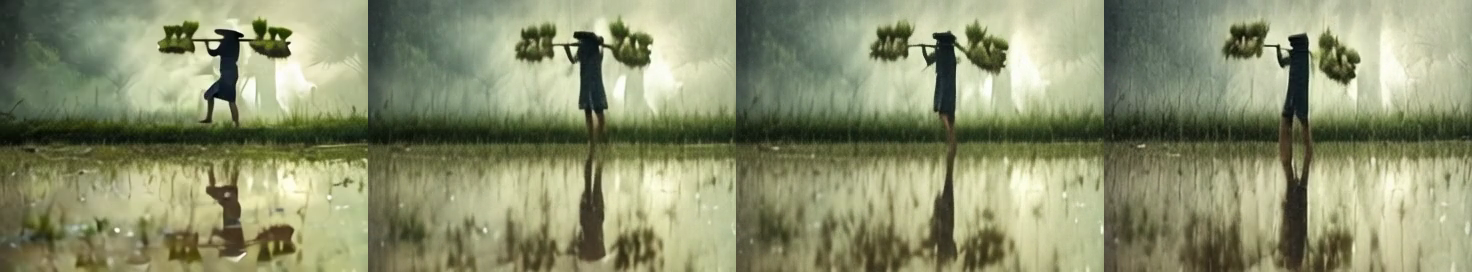} \\
\rotatebox[origin=l]{90}{~~~+VCD} & \includegraphics[width=0.93\linewidth]{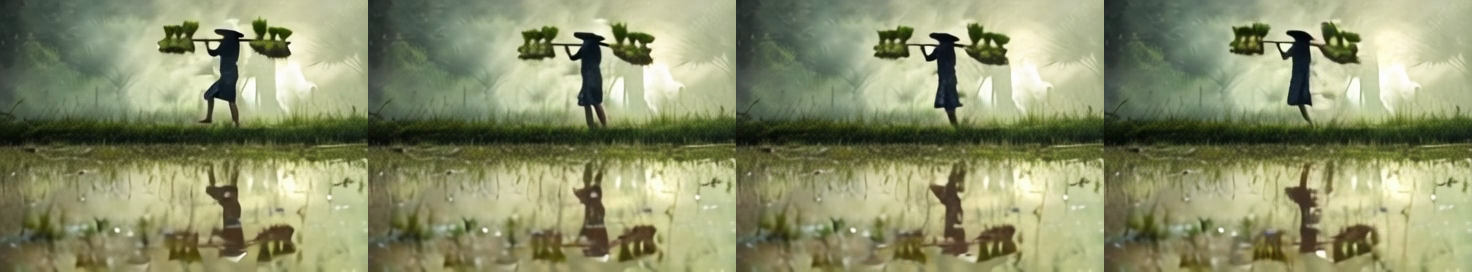} \\
 & \textit{``a woman carrying a bundle of plants over their head''} \\
 \\
 \rotatebox[origin=l]{90}{~~~~~Wan} & \includegraphics[width=0.93\linewidth]{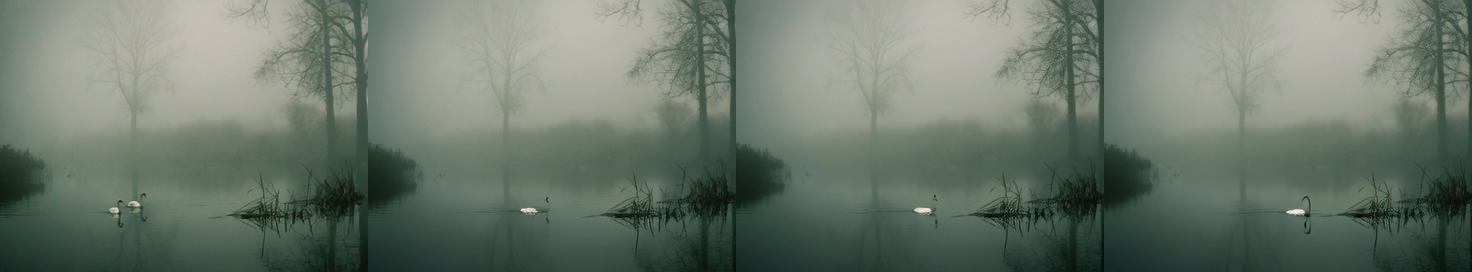} \\
\rotatebox[origin=l]{90}{~~~+V-JEPA} & \includegraphics[width=0.93\linewidth]{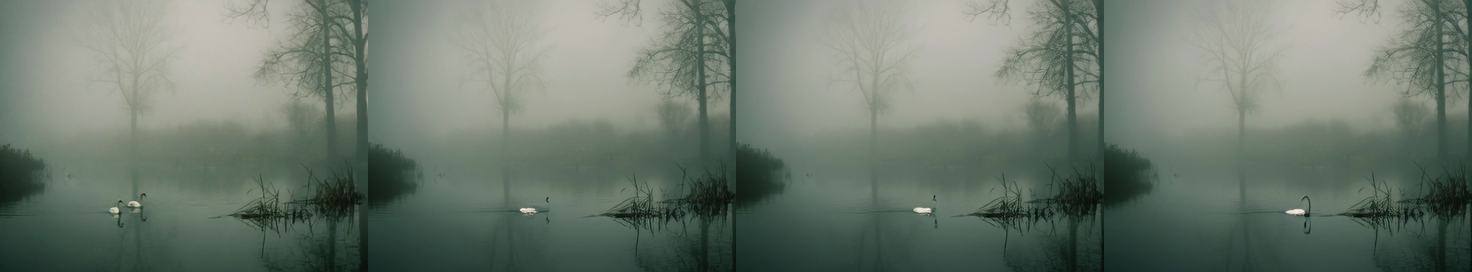} \\
\rotatebox[origin=l]{90}{~~~+VCD} & \includegraphics[width=0.93\linewidth]{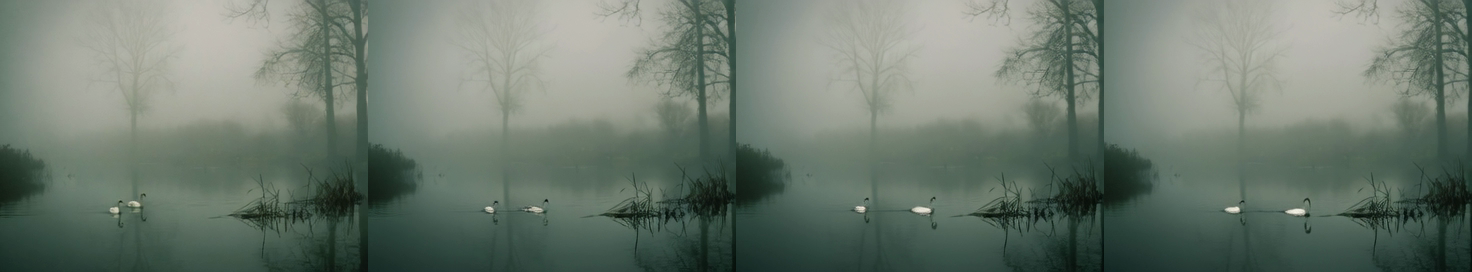} \\
 & \textit{``two swans swimming on a lake in the fog''} \\
\end{tabular}
\caption{
Results of video generation with VBench-I2V.
}
\label{fig:results-opensora-vbench_i2v}
\vspace*{-3mm}
\end{figure}

\begin{figure}[t]
\centering
\footnotesize
\tabcolsep=.5mm
\begin{tabular}{cc}
\rotatebox[origin=l]{90}{~~Open-Sora} & \includegraphics[width=0.93\linewidth]{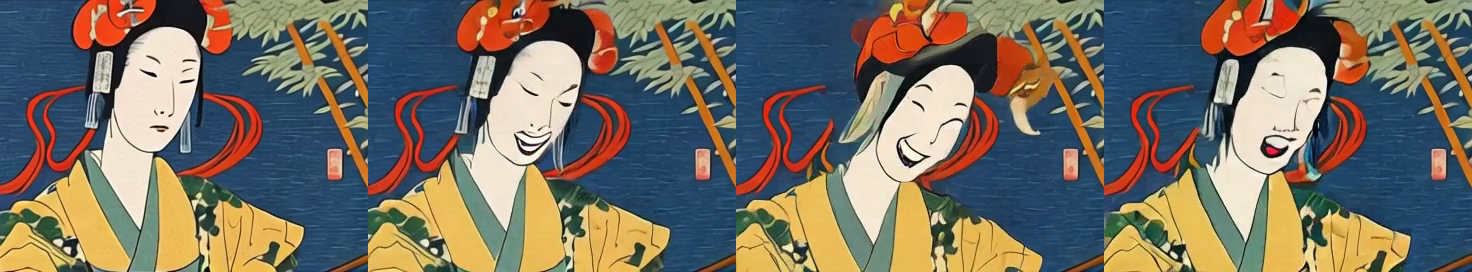} \\
 \rotatebox[origin=l]{90}{~~~+V-JEPA} & \includegraphics[width=0.93\linewidth]{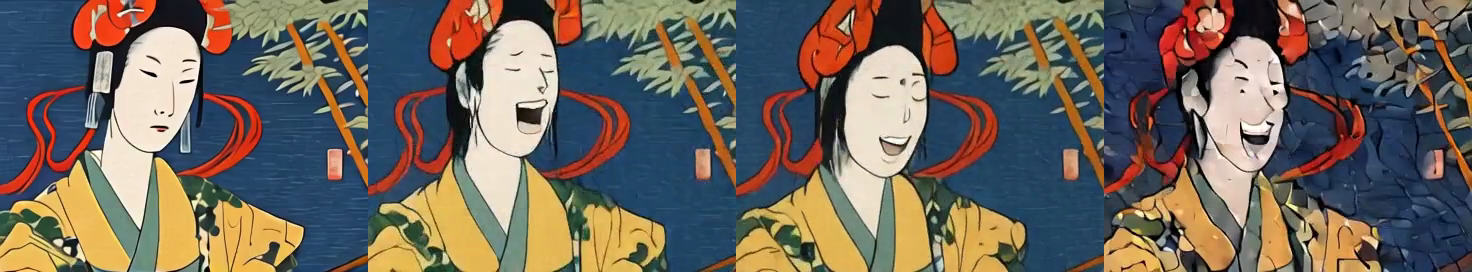} \\
 \rotatebox[origin=l]{90}{~~~+VCD} & \includegraphics[width=0.93\linewidth]{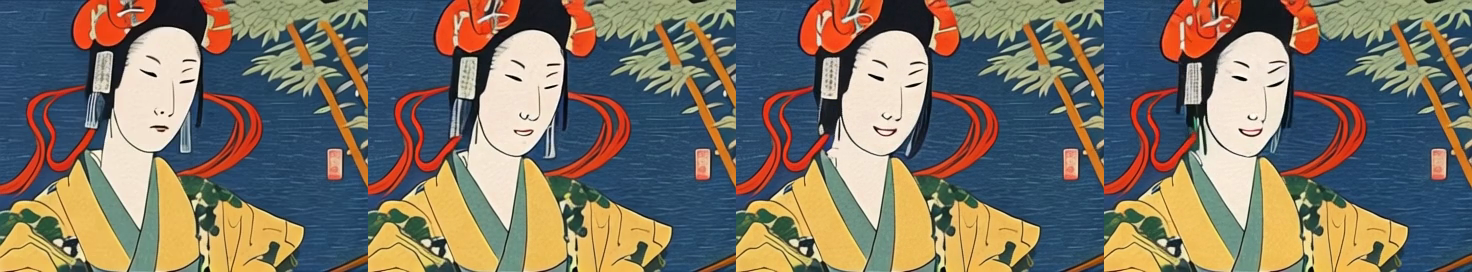} \\
  & \textit{``a person is singing happily''} \\
  \\
  \rotatebox[origin=l]{90}{~~~~~Wan} & \includegraphics[width=0.93\linewidth]{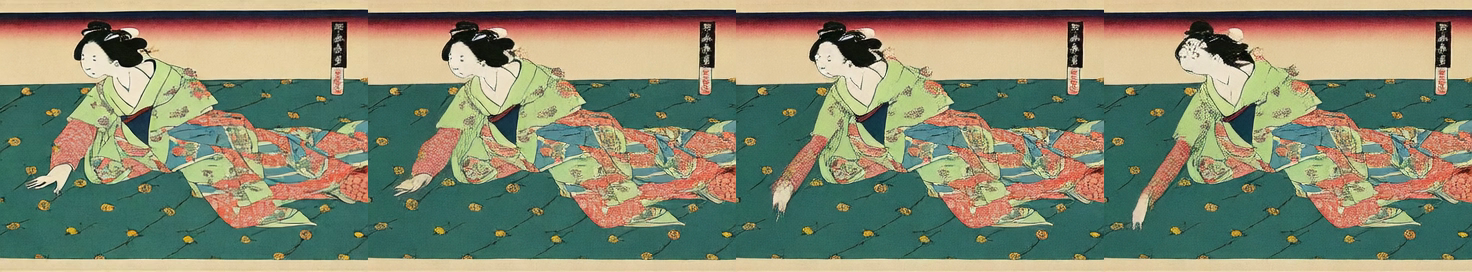} \\
 \rotatebox[origin=l]{90}{~~~+V-JEPA} & \includegraphics[width=0.93\linewidth]{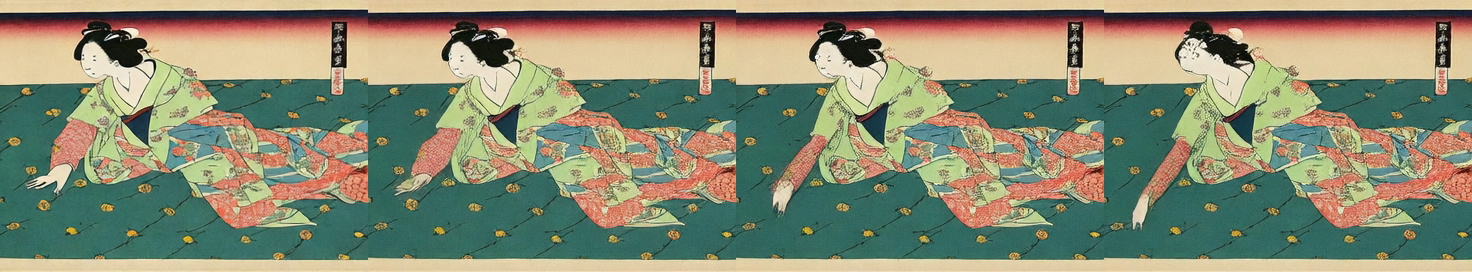} \\
 \rotatebox[origin=l]{90}{~~~+VCD} & \includegraphics[width=0.93\linewidth]{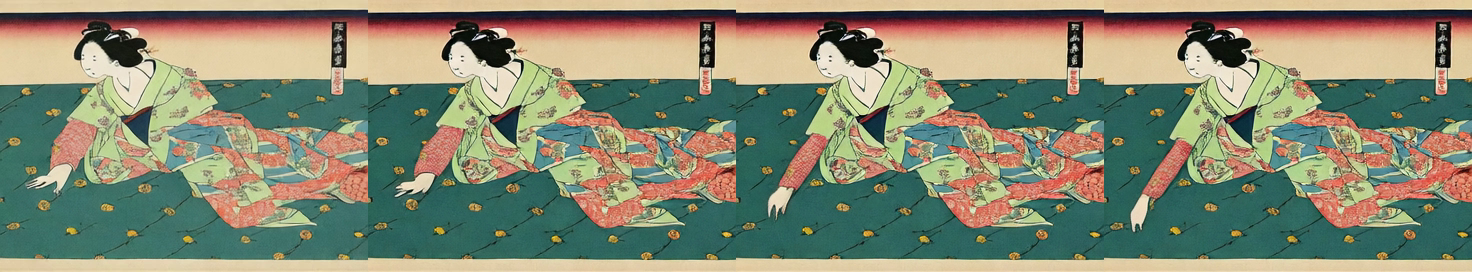} \\
  & \textit{``a person is reaching forward over a floral-patterned surface''} \\
\end{tabular}
\caption{
Results of video generation with AI-ArtBench.
}
\label{fig:results-opensora-ai_artbench}
\end{figure}

\section{Experiments}
\subsection{Experimental Setting}
\paragraph{Datasets}
For our experiments, we employed the following three datasets: I2V-Bench~\cite{ren2024consisti2v}, VBench-I2V~\cite{huang2023vbench,huang2024vbench++}, and AI-ArtBench dataset~\cite{silva2024artbrainexplainableendtoendtoolkit}.
I2V-Bench consists of 2,951 high-quality videos accompanied by corresponding captions.
We randomly divided the text-video pairs into train and evaluation sets with 100 and 2,851 samples, respectively.
On the other hand, VBench-I2V consists of 355 images, each image associated with one or more captions.
VBench-I2V provides an evaluation benchmark for I2V models.
Since it requires all the images and captions in the VBench-I2V dataset, we did not use them for fine-tuning.
Instead, we used the 100 I2V-Bench videos for fine-tuning when evaluating models on VBench-I2V.
For evaluation in more various domains, we employed AI-ArtBench, an AI-generated art dataset that contains more than 180,000 images with multiple style subsets.
We specifically used the ``ukiyoe'' subset, containing about 12,000 images, a unique domain in standard datasets for video generation models.
We randomly selected 25 and 100 images for fine-tuning and evaluation, respectively.
Note that we manually annotated the prompts for the images of AI-ArtBench used in our experiment because the original dataset does not provide text prompts.

\paragraph{Models Details}
We employed two state-of-the-art video generation models, Open-Sora~\cite{opensora} and Wan2.1-1.3B-I2V (Wan)~\cite{wan2025}, as the baseline models.
We generated videos with 51 frames at a resolution of $368\times 272$ for Open-Sora, and 25 frames at a resolution of $368\times 272$ for Wan.

For comparative methods, we chose VADER, where V-JEPA~\cite{bardes2024revisiting} was employed as a reward function, because it focused on enhancing temporal consistency.
We did not employ other reward functions, such as HPS and PickScore, because they are not suitable for I2V generation as shown in Fig.~\ref{fig:unconsistent} and Fig.~\ref{fig:appendix-results-others}.
For V-JEPA, we employed ViT-H/16~\cite{dosovitskiy2021an} as the backbone architecture.

For fine-tuning the baseline models, we employed AdamW optimizer~\cite{loshchilov2018decoupled} with a constant learning rate $2\times 10^{-4}$.
The denoising process spans 30 steps for generating samples.
Note that one fine-tuning took less than two days.
To reduce memory consumption of fine-tunings, we employed LoRA~\cite{hu2022lora}, truncated backpropagation~\cite{tallec2017unbiasingtruncatedbackpropagationtime}, and subsampling frames for Open-Sora, and TeaCache~\cite{liu2024timestep} for Wan.
For truncated backpropagation, we only backpropagated through the final denoising step.

\paragraph{Evaluation Metrics}
For evaluation metrics, we employed two comprehensive benchmarks, namely VBench-I2V~\cite{huang2023vbench,huang2024vbench++} and VideoScore~\cite{he2024videoscore}.
VBench-I2V measures I2V generation quality across ten evaluation dimensions (I2V Subject, I2V Background, Camera Motion, Subject Consistency, Background Consistency, Motion Smoothness, Dynamic Degree, Aesthetic Quality, Imaging Quality, and Temporal Flickering).
Among these metrics, I2V Subject, I2V Background, Subject Consistency, Background Consistency, and Temporal Flickering measure temporal consistency.
Specifically, I2V Subject and I2V Background measure temporal consistency relative to the conditioning image, while the others measure temporal consistency across whole generated frames.
Camera Motion, Motion Smoothness, and Dynamic Degree focus on the extent or smoothness of motion, whereas Aesthetic Quality and Imaging Quality assess the overall beauty of the generated frames.
Since VBench-I2V does not provide a video-text alignment metric, we evaluated it by calculating the ViCLIP~\cite{wang2023internvid} feature similarity between a generated video and a conditioning text prompt.

On the other hand, VideoScore~\cite{he2024videoscore} evaluates generated videos from five aspects, including Visual Quality, Temporal Consistency, Dynamic Degree, Text-Video Alignment, and Factual Consistency, using the fine-tuned MantisIdefics2-8B~\cite{Jiang2024MANTISIM} with a human-annotated generated videos dataset of the above five metrics.
Note that Temporal Consistency in VideoScore evaluates the consistency through a whole video sequence.

In all experiments, we generated five videos for each pair of conditioning images and text prompts with different random seeds to capture model variability and ensure robust evaluation.
See Appendix~\ref{sec:appendix-details-settings} for the details of the evaluation metrics.

\subsection{Experimental Results}
\label{subsec:results}
\paragraph{Qualitative Results}
Figures~\ref{fig:results-opensora-i2v_bench}, \ref{fig:results-opensora-vbench_i2v}, and \ref{fig:results-opensora-ai_artbench} show the samples generated by the baseline models and fine-tuned models along with V-JEPA and the proposed method.
Note that +V-JEPA and +VCD in the figures refer to the fine-tuned models with each reward function.

Overall, while the baseline models exhibited temporally inconsistent videos, the fine-tuned models (+V-JEPA and +VCD) demonstrated improved temporal consistency (\eg, the top part of Fig.~\ref{fig:results-opensora-vbench_i2v}).
Furthermore, we also observed the improvement in temporal consistency by +V-JEPA over the baseline models in Fig.~\ref{fig:results-opensora-i2v_bench}.
These results support VADER~\cite{prabhudesai2024videodiffusionalignmentreward}'s qualitative demonstration, which showed that fine-tuning a video diffusion model with V-JEPA enhances temporal consistency.

However, +V-JEPA sometimes retained temporal inconsistency in the generated videos.
For example, as shown in the top part of Fig.~\ref{fig:results-opensora-i2v_bench}, +V-JEPA generated drastically different frames relative to the conditioning image, same as the baseline model.
This might be because V-JEPA focuses on enhancing overall temporal consistency by utilizing the global feature from the entire video sequence, thereby overlooking the preservation of temporal consistency relative to a conditioning image.
In contrast, VCD focuses on enhancing temporal consistency relative to a conditioning image.
Therefore, as shown in the top part of Fig.~\ref{fig:results-opensora-i2v_bench}, the generated samples by the fine-tuned model with VCD did not show such drastic changes.
On the other hand, as shown in Fig.~\ref{fig:results-opensora-ai_artbench}, while the human faces and the objects are distorted in the samples generated by the baseline models and +V-JEPA, +VCD remained faithful to the conditioning images throughout the generated frames.
We provide additional results in Appendix~\ref{sec:appendix-additional-results}.

\begin{figure*}[t]
\centering
\includegraphics[width=1\linewidth]{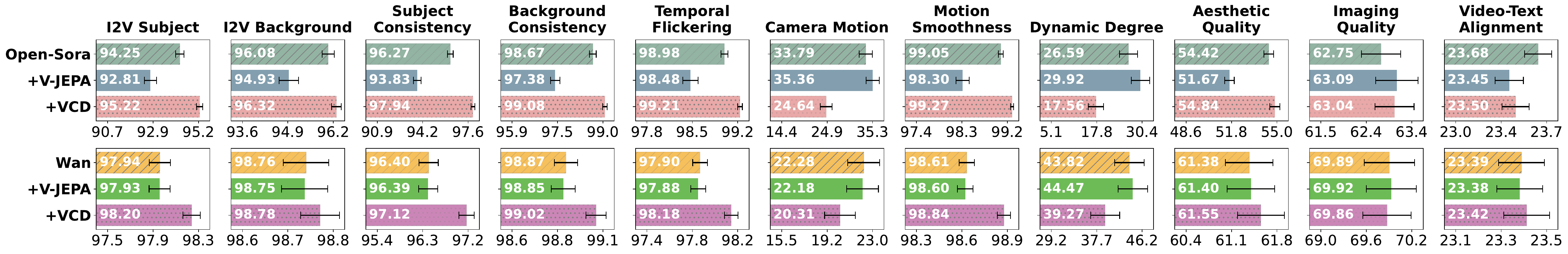}
\caption{VBench-I2V and Video-Text Alignment of Open-Sora and its fine-tuned models [\%].
The values in each bar and the error bars represent the means and 95\% confidence intervals of five runs, respectively.}
\label{fig:quantitative-vbench_i2v}
\end{figure*}

\begin{figure*}[t]
  \begin{minipage}[b]{0.48\linewidth}
    \centering
    \includegraphics[width=1\linewidth]{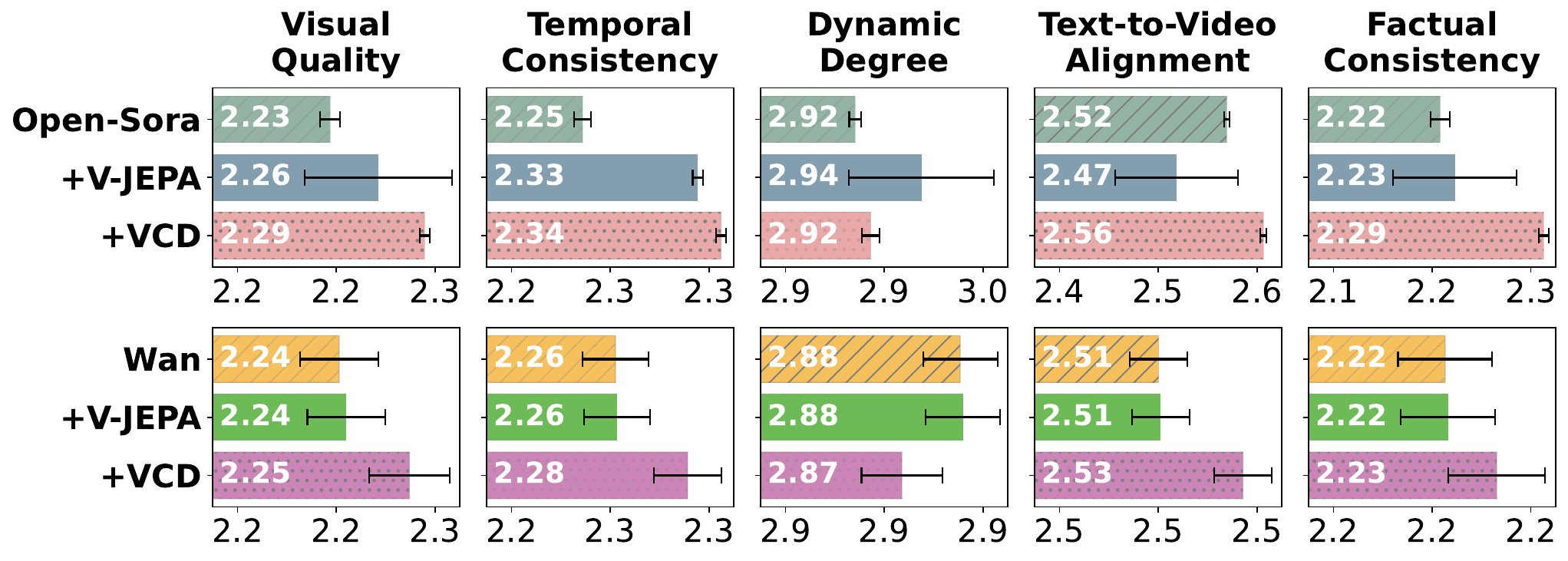}
    \subcaption{I2V-Bench}
  \end{minipage}
  \begin{minipage}[b]{0.48\linewidth}
    \centering
    \includegraphics[width=1\linewidth]{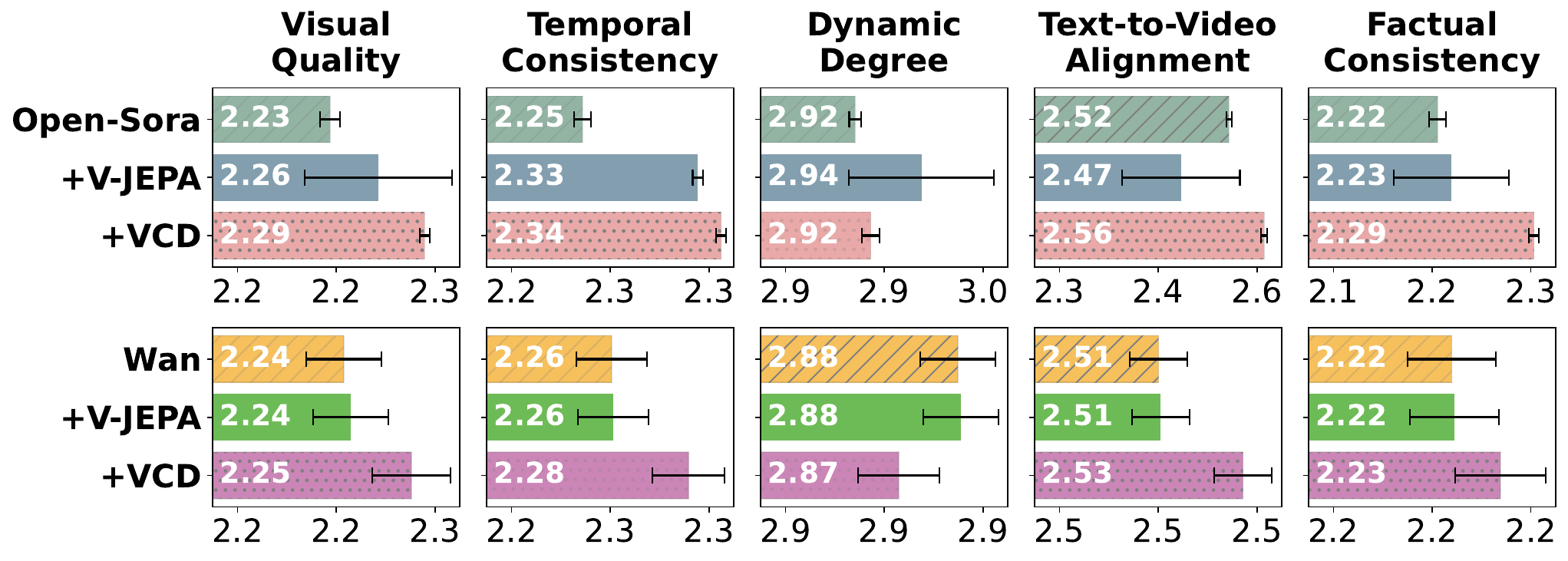}
    \subcaption{AI-ArtBench}
  \end{minipage}
  \caption{VideoScore of baseline models and their fine-tuned models in I2V-Bench and AI-ArtBench.
  A higher score indicates relatively better performance.}
  \label{fig:quantitative-videoscore}
\end{figure*}

\paragraph{Quantitative Results}
Figure~\ref{fig:quantitative-vbench_i2v} illustrates the VBench-I2V and the Video-Text Alignment scores on the VBench-I2V dataset, and Fig.~\ref{fig:quantitative-videoscore} presents the VideoScore results on the I2V-Bench and AI-ArtBench datasets.
We also summarized them in Table~\ref{tab:quantitative-vbench_i2v} and Table~\ref{tab:quantitative-videoscore} in Appendix~\ref{sec:appendix-additional-results}.

As shown in Fig.~\ref{fig:quantitative-videoscore}, a comparison between the baseline models and +V-JEPA on the Temporal Consistency and Factual Consistency scores of VideoScore provides evidence to imply their contribution to improving temporal consistency of generated videos.
Furthermore, as shown in Fig.~\ref{fig:quantitative-vbench_i2v} and \ref{fig:quantitative-videoscore}, +V-JEPA showed improvements in the Dynamic Degree score compared to the baseline models.
Moreover, +V-JEPA yielded better results in the appearance metric (\ie, Imaging Quality in the VBench-I2V scores), as reported in VADER~\cite{prabhudesai2024videodiffusionalignmentreward}.

However, as shown in Fig.~\ref{fig:quantitative-vbench_i2v}, the results on the VBench-I2V dataset indicate that incorporating V-JEPA did not lead to statistically significant improvements with $p<0.05$ on the other metrics.
In particular, +V-JEPA showed statistically significant deterioration compared to Open-Sora on the I2V Subject, Subject Consistency, Background Consistency, and Temporal Flickering scores, which are related to the temporal consistency of generated videos.
These results support our previously mentioned hypothesis regarding the limitations of utilizing V-JEPA as a reward function for fine-tuning I2V generation models.

On the other hand, in Fig.~\ref{fig:quantitative-vbench_i2v} and Fig.~\ref{fig:quantitative-videoscore}, +VCD achieved the highest performance on the above five temporal consistency metrics in the VBench-I2V scores and Temporal Consistency and Factual Consistency scores in VideoScore.
Additionally, our approach yielded better or comparable scores than the baseline models across various metrics, including Aesthetic Quality and Imaging Quality scores (as measured by the VBench-I2V scores), as well as Visual Quality and Text-to-Video Alignment scores (as measured by VideoScore).
These results indicate that fine-tuning with VCD effectively enhances temporal consistency without compromising other qualities, and it does not depend on the baseline model's performance or datasets.
However, in terms of dynamic-related metrics (\ie, Dynamic Degree and Camera Motion), +VCD yielded lower scores compared to the baseline models and +V-JEPA.
This limitation may be due to VCD's firm adherence to the conditioning frame, which may constrain its ability to generate large or exaggerated motion.

\begin{figure*}
    \centering
    \includegraphics[width=1.0\linewidth]{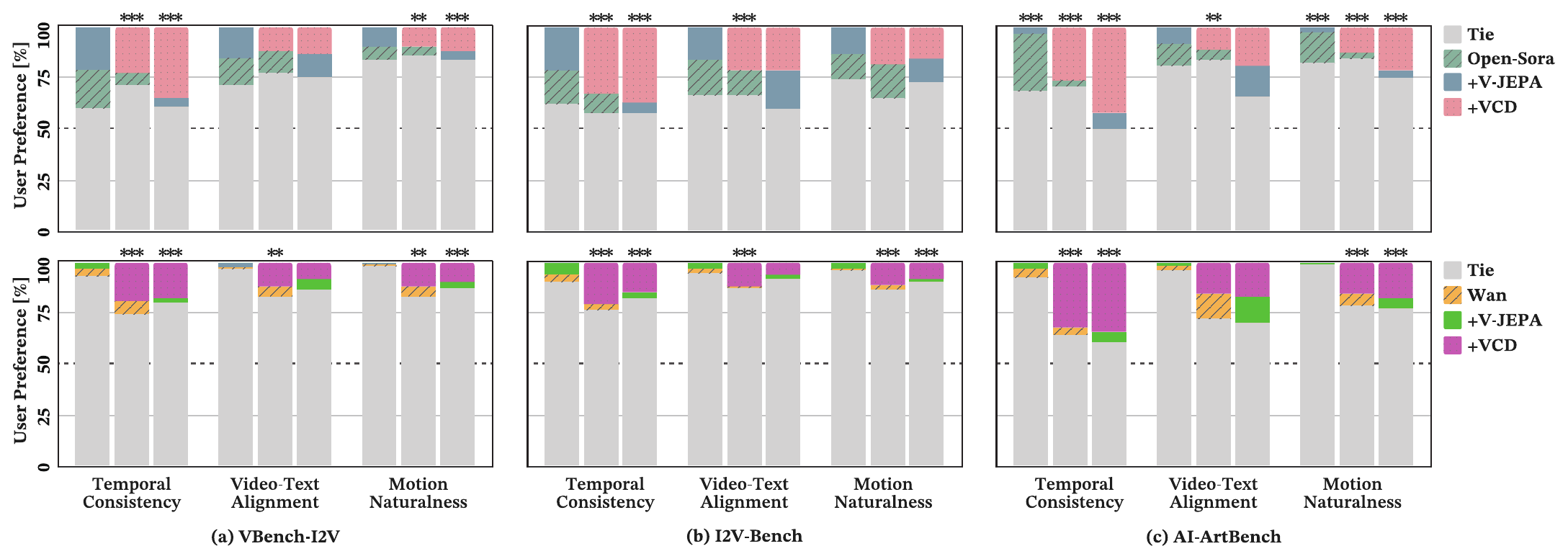}
    \caption{
        Human evaluation results.
        ``Tie'' indicates that annotators evaluated two videos as comparable.
        The three bars in each metric represent, from left to right, a baseline model (Open-Sora or Wan) vs. +V-JEPA, a baseline model vs. +VCD, and +V-JEPA vs. +VCD, respectively.
        \textasteriskcentered\textasteriskcentered\ and \textasteriskcentered\textasteriskcentered\textasteriskcentered\ above bars indicates that the result showed statistically significant improvement with $0.001 \leq p < 0.005$ and $p < 0.001$ of the t-test, respectively.
        Results without \textasteriskcentered\textasteriskcentered\ or \textasteriskcentered\textasteriskcentered\textasteriskcentered\ did not show statistically significant improvement.
    }
    \label{fig:results-human-evaluation}
\end{figure*}

\paragraph{Human Evaluation Results}
To assess the effectiveness of each method from the perspective of human perception, we conducted a human evaluation study.
From generated videos with I2V-Bench, VBench-I2V, and AI-ArtBench images, we randomly selected 30 videos per dataset, resulting in 90 videos in total for each model.
We collected 15 human evaluators, who were individuals with prior experience in video quality assessment, but were not themselves researchers in computer vision.
Evaluators were presented each with the conditioning image, text prompt, and pairs of videos generated by any two of the baseline models, +V-JEPA, or +VCD.
We provide a screenshot of user interface in Fig.~\ref{fig:appendix-ui} in Appendix~\ref{sec:appendix-details-settings}.
They were tasked to judge which video was better or whether they were equivalent in terms of Temporal Consistency, Video-Text Alignment, and Motion Naturalness.
Note that we encouraged evaluators to independently assess the video's motion naturalness, without considering the conditioning image and text prompt, to ensure unbiased evaluation from other aspects.
We provide details of human evaluation settings in Appendix~\ref{sec:appendix-details-settings}.

Figure~\ref{fig:results-human-evaluation} shows the human evaluation results.
We also summarized them in Table~\ref{tab:quantitative-human} in Appendix~\ref{sec:appendix-additional-results}.
We performed a t-test to evaluate whether the observed differences were statistically significant.
In all metrics and in all datasets, +VCD showed better scores than the baseline models and +V-JEPA.
In particular, +VCD showed statistically significant improvements in Temporal Consistency across all datasets, with $p < 0.001$.
Moreover, +VCD outperformed the baseline models and +V-JEPA in Video-Text Alignment and Motion Naturalness in all datasets.
These results indicate that +VCD generated temporally consistent videos with natural motion and better video-text alignment compared to the baseline models and +V-JEPA.
+V-JEPA showed better or comparable results in Temporal Consistency in I2V-Bench and VBench-I2V compared to Open-Sora.
However, in AI-ArtBench, it showed statistically significant decreases compared to Open-Sora and lower results compared to Wan, likely due to the dependence on its training dataset.
These results imply that fine-tuning a video diffusion model with V-JEPA may cause worse results in unseen domains, in this case, artistic images.
In contrast, VCD utilizes only the shallow layers of VGG19 and is not heavily dependent on its training dataset.
As a result, +VCD demonstrated significantly greater robustness on the AI-ArtBench dataset compared to +V-JEPA.

\subsection{Ablation Study}

\begin{table}[t]
  \centering
  \tabcolsep=0.5mm
  \caption{Parts of VBench-I2V for an ablation study about the individual contribution of amplitude and phase components of the proposed method [\%]. Amp. refers to amplitude.}
  \label{tab:quantitative-vbench_i2v-ablation}
  \begin{tabular}{lccc}
    \toprule
     & \textbf{I2V} & \textbf{I2V} & \textbf{Temporal}\\
     & \textbf{Subject} & \textbf{Background} & \textbf{Flickering}  \\
    \midrule
    Open-Sora & 94.25\scriptsize{$\pm$0.21} & 96.08\scriptsize{$\pm$0.18} & 98.98\scriptsize{$\pm$0.05} \\
    \midrule
    +Amp. & \textbf{95.67}\scriptsize{$\pm$0.12} & \textbf{96.40}\scriptsize{$\pm$0.11} & 97.76\scriptsize{$\pm$0.11} \\
    \midrule
    +Phase & 91.88\scriptsize{$\pm$0.34} & 95.65\scriptsize{$\pm$0.26} & \textbf{99.40}\scriptsize{$\pm$0.02} \\
    \midrule
    +VCD \scriptsize{(+Amp. \& +Phase)} & \underline{95.22}\scriptsize{$\pm$0.16} & \underline{96.32}\scriptsize{$\pm$0.14} & \underline{99.21}\scriptsize{$\pm$0.04} \\
    \bottomrule
  \end{tabular}
  \vspace{-3mm}
\end{table}

For generating a temporally consistent video, it is essential to preserve both global and local attributes over time.
In Section~\ref{subsec:vcd}, we explained how amplitude and phase components contribute to preserving global and local attributes, respectively.
To validate this, we trained Open-Sora with the first term of Eq.~\ref{eq:vcd} (amplitude) and the second term of Eq.~\ref{eq:vcd} (phase).
Table~\ref{tab:quantitative-vbench_i2v-ablation} shows the parts of the VBench-I2V results.
I2V Subject and I2V Background evaluate temporal consistency of global attributes.
On the other hand, Temporal Flickering evaluates temporal consistency of local attributes.
+Amp. outperformed on I2V Subject and I2V Background, indicating that the amplitude components contribute to the preservation of global attributes.
In contrast, +Phase showed a higher Temporal Flickering, suggesting that phase components contribute to preserving local attributes.
Since VCD combines both amplitude and phase components, its performance on each individual metric is lower than +Amp. or +Phase alone.
However, benefiting from both contributions, VCD achieved superior results across all metrics compared with the baseline model.
These experimental results support the claims described in Section~\ref{subsec:vcd} regarding the individual contribution of the amplitude and phase components.
We provide other ablation studies in Appendix~\ref{sec:appendix-ablation}.

\section{Conclusion}
In this paper, we proposed Video Consistency Distance (VCD) to enhance temporal consistency in I2V generation.
We experimentally showed that fine-tuning a model with VCD enhances temporal consistency relative to a conditioning image without degrading other performance.
A limitation is that a model fine-tuned with VCD struggles to generate a video that contains large motions.
Future work will focus on handling motion strength by employing adaptive temporal weight, for example, by employing Multimodal Large Language Models (\eg, PLLaVA~\cite{xu2024pllava}) to estimate motion strength from the prompt and adapt to temporal weight.

{
    \small
    \bibliographystyle{ieeenat_fullname}
    \bibliography{egbib}
}
\clearpage
\appendix

\section{Other Reward Functions}
\label{sec:appendix-other-reward-functions}
\begin{figure}[t]
\centering
\footnotesize
\tabcolsep=.4mm
\begin{tabular}{cc}
\rotatebox[origin=l]{90}{~+Aesthetic} & \includegraphics[width=0.95\linewidth]{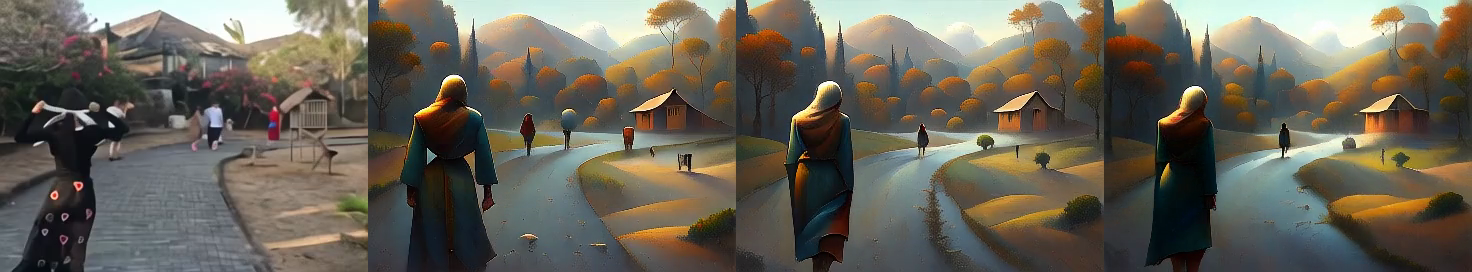}\\
\rotatebox[origin=l]{90}{+PickScore} & \includegraphics[width=0.95\linewidth]{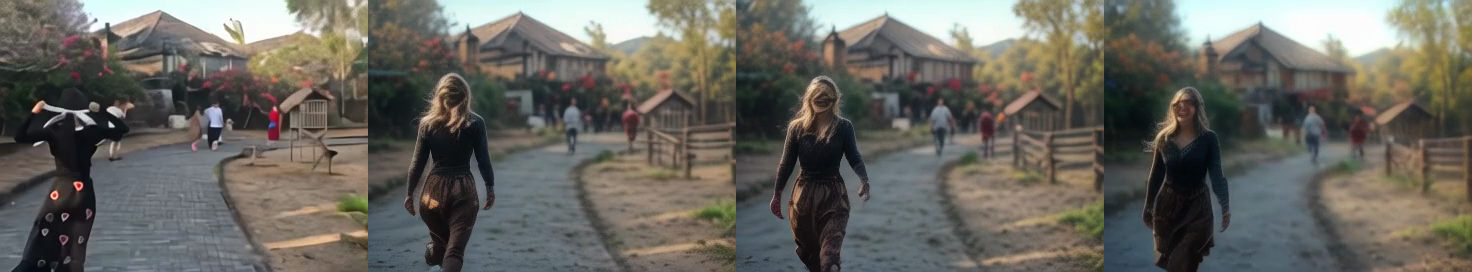} \\
\rotatebox[origin=l]{90}{~~~~+HPS} & \includegraphics[width=0.95\linewidth]{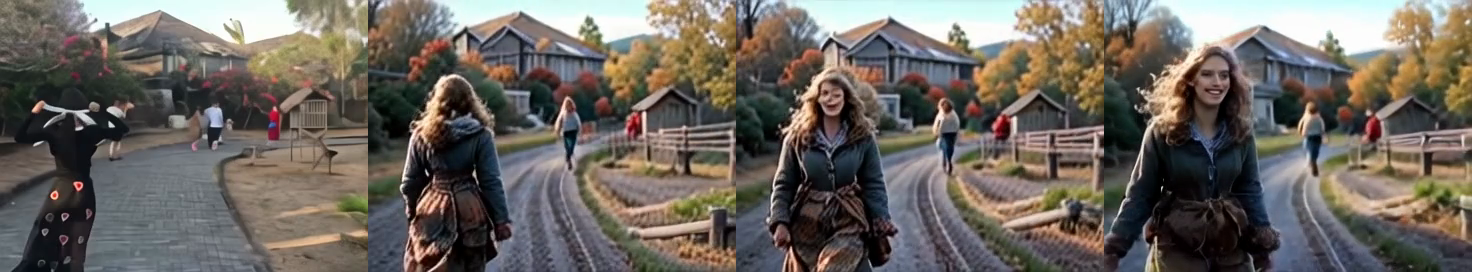} \\
 & \textit{``a woman was walking happily on the farm''} \\
 \\
 \rotatebox[origin=l]{90}{~+Aesthetic} & \includegraphics[width=0.95\linewidth]{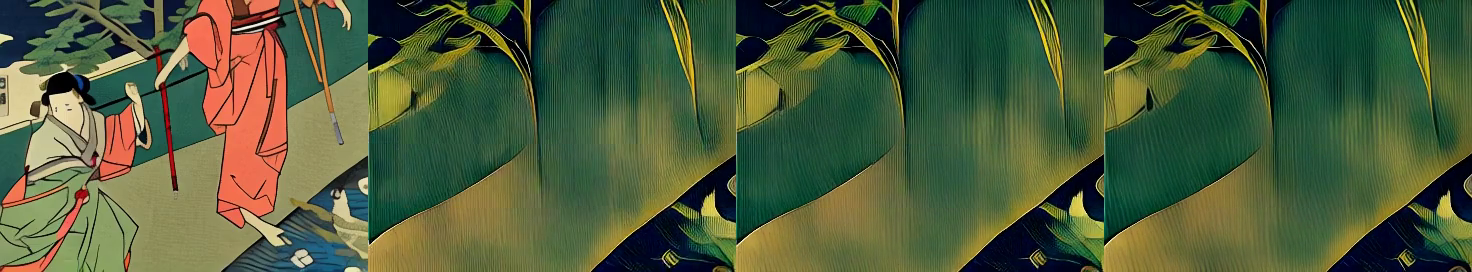}\\
\rotatebox[origin=l]{90}{+PickScore} & \includegraphics[width=0.95\linewidth]{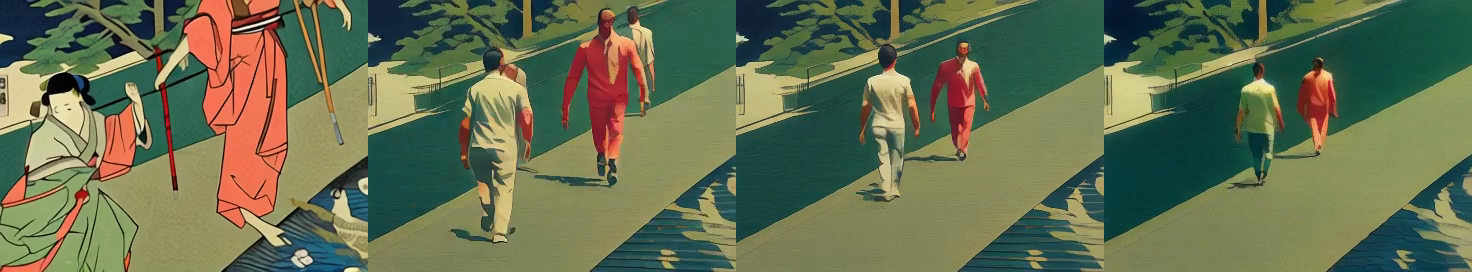} \\
\rotatebox[origin=l]{90}{~~~~+HPS} & \includegraphics[width=0.95\linewidth]{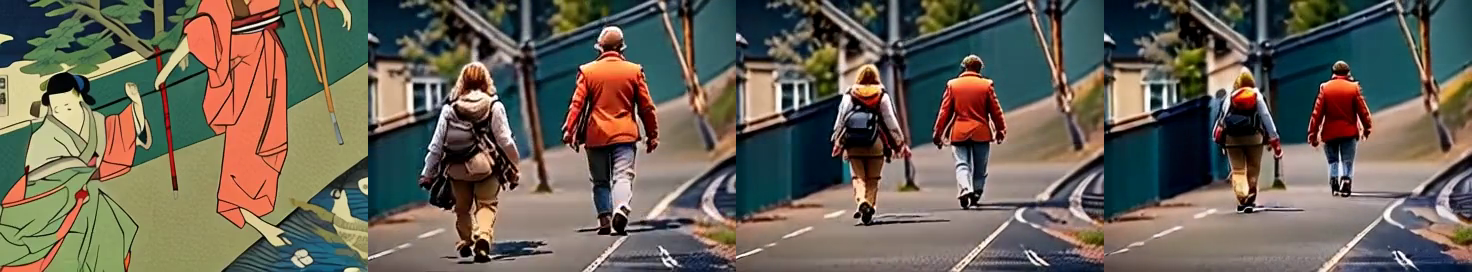} \\
 & \textit{``two people are walking from left to right.''} \\
\end{tabular}
\caption{
Results of videos generated by fine-tuning models with Aesthetic~\cite{aesthetic}, PickScore, and HPS, respectively.
We provide text prompts below the figures.
}
\label{fig:appendix-results-others}
\end{figure}

VADER employed HPS~\cite{wu2023human}, PickScore~\cite{kirstain2023pickapicopendatasetuser}, and LAION Aesthetic predictor (Aesthetic)~\cite{aesthetic}.
In our preliminary experiments, we found that the fine-tuned models with these functions generated drastically different frames relative to a conditioning image, as shown in Fig.~\ref{fig:appendix-results-others}.

\begin{figure*}[t]
\centering
\includegraphics[width=0.95\linewidth]{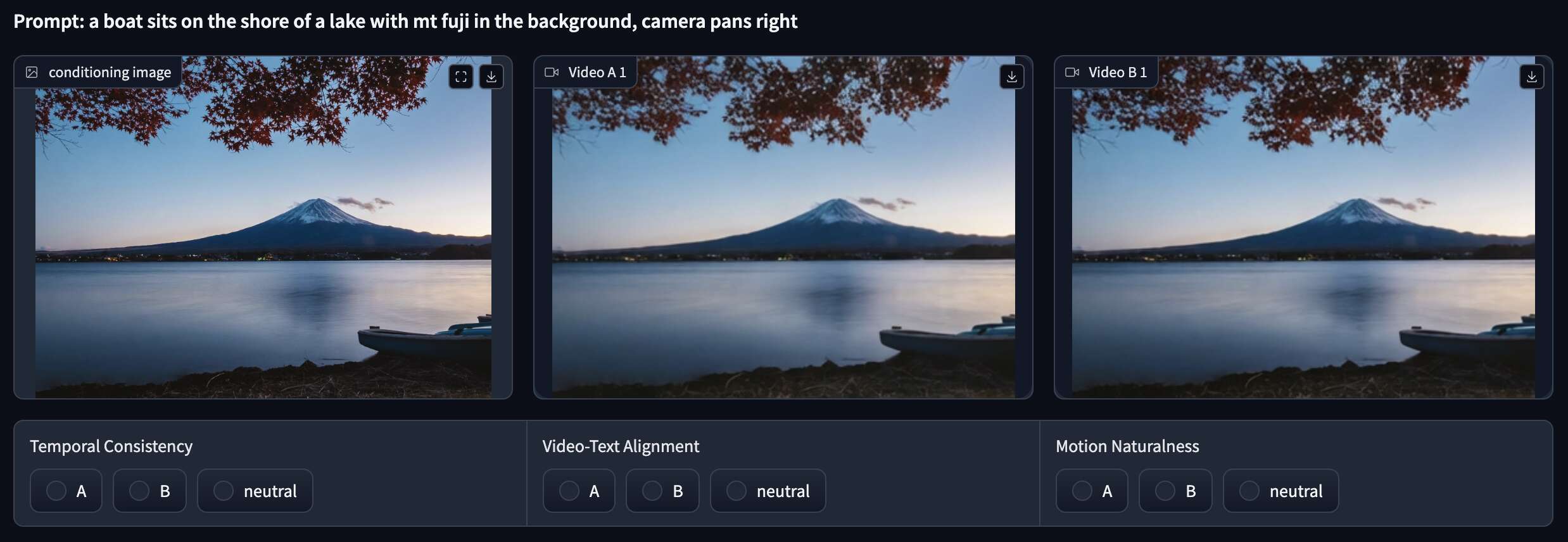}
\caption{
A screenshot of our user interface for human evaluation.
}
\label{fig:appendix-ui}
\end{figure*}

\begin{figure}[t]
\centering
\footnotesize
\tabcolsep=.4mm
\begin{tabular}{cc}
\rotatebox[origin=l]{90}{~~Open-Sora} & \includegraphics[width=0.95\linewidth]{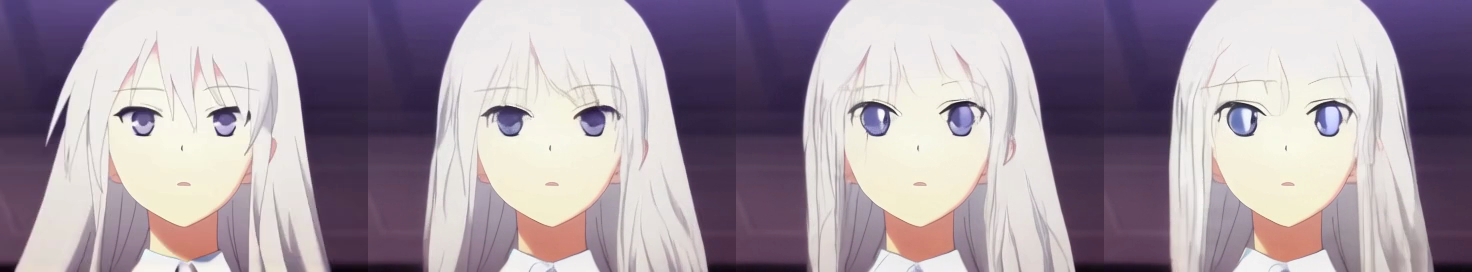}\\
\rotatebox[origin=l]{90}{~~~+V-JEPA} & \includegraphics[width=0.95\linewidth]{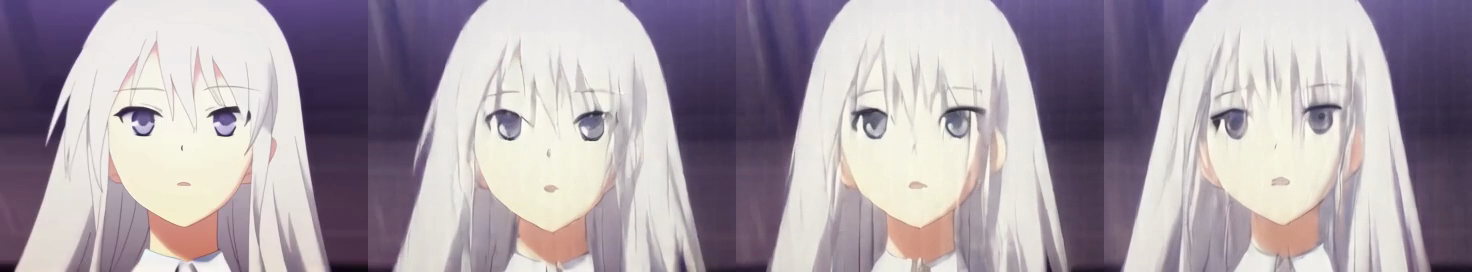} \\
\rotatebox[origin=l]{90}{~~~+VCD} & \includegraphics[width=0.95\linewidth]{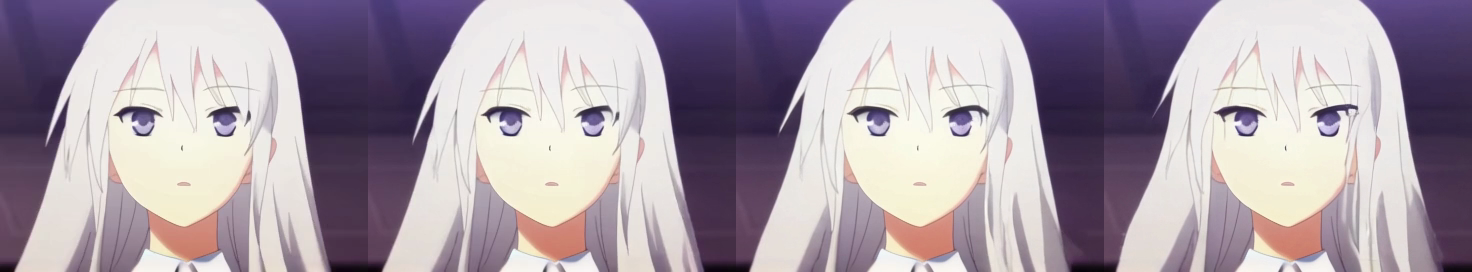} \\
 & \textit{``girl's hair swing with the wind''} \\
 \\
 \rotatebox[origin=l]{90}{~~~~~Wan} & \includegraphics[width=0.95\linewidth]{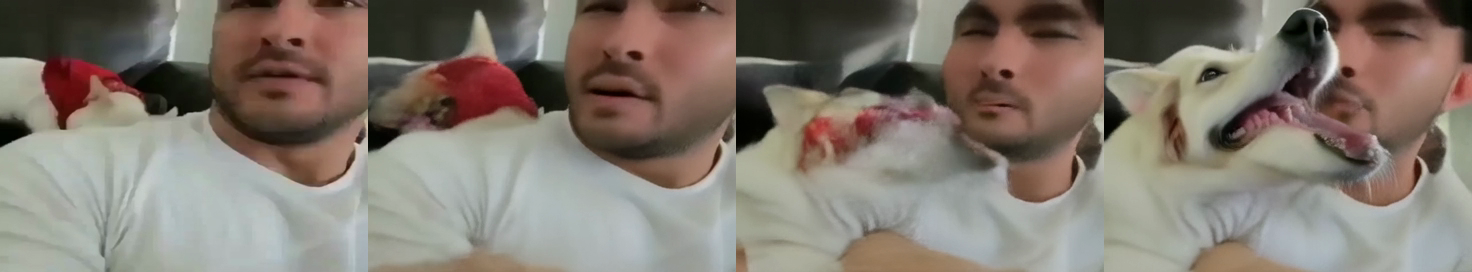}\\
\rotatebox[origin=l]{90}{~~~+V-JEPA} & \includegraphics[width=0.95\linewidth]{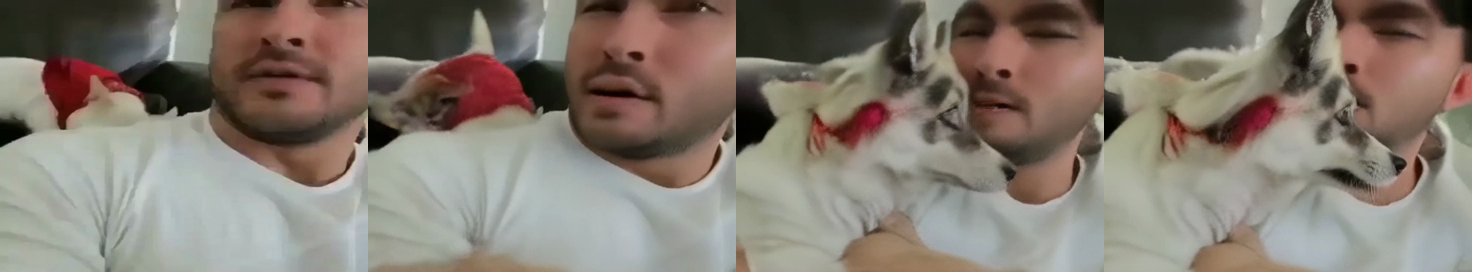} \\
\rotatebox[origin=l]{90}{~~~+VCD} & \includegraphics[width=0.95\linewidth]{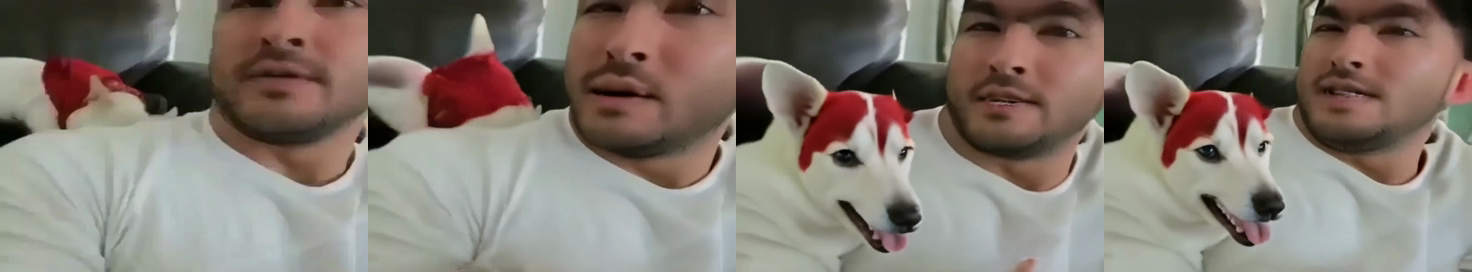} \\
 & \textit{``a white puppy is interacting with the male owner on the sofa''} \\
\end{tabular}
\caption{
Results of video generation with I2V-Bench.
We provide text prompts below the figures.
}
\label{fig:appendix-results-opensora-i2v_bench}
\end{figure}

\begin{figure}[t]
\centering
\footnotesize
\tabcolsep=.4mm
\begin{tabular}{cc}
 \rotatebox[origin=l]{90}{~~Open-Sora} & \includegraphics[width=0.95\linewidth]{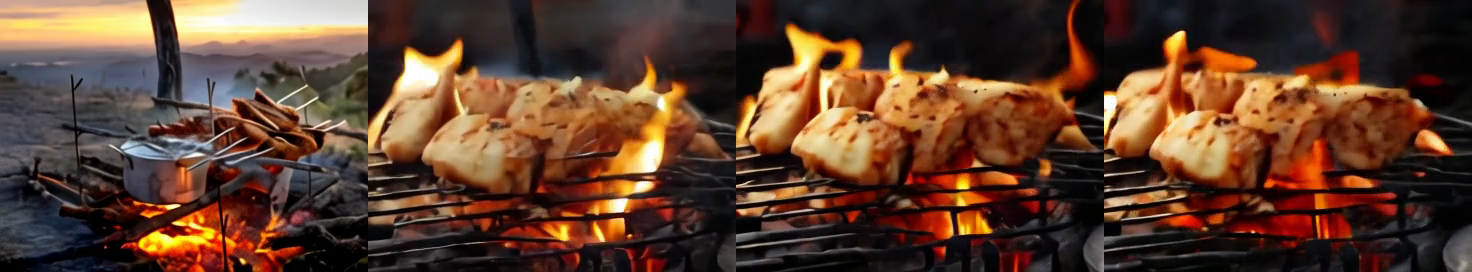} \\
 \rotatebox[origin=l]{90}{~~~+V-JEPA} & \includegraphics[width=0.95\linewidth]{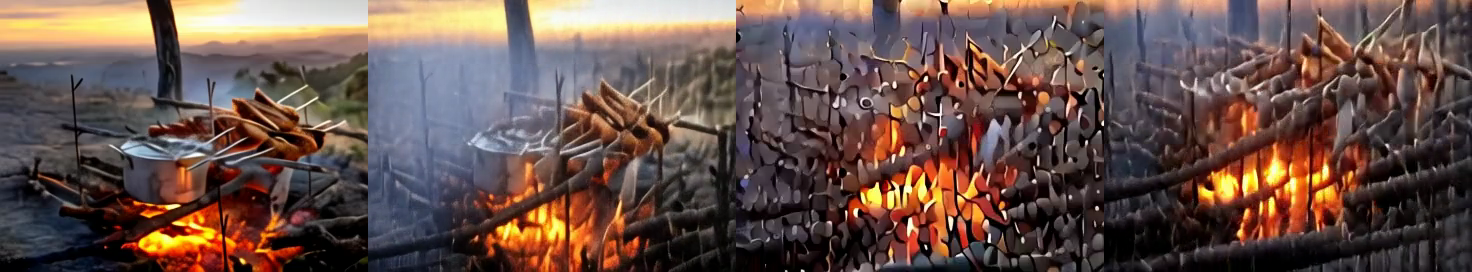} \\
 \rotatebox[origin=l]{90}{~~~+VCD} & \includegraphics[width=0.95\linewidth]{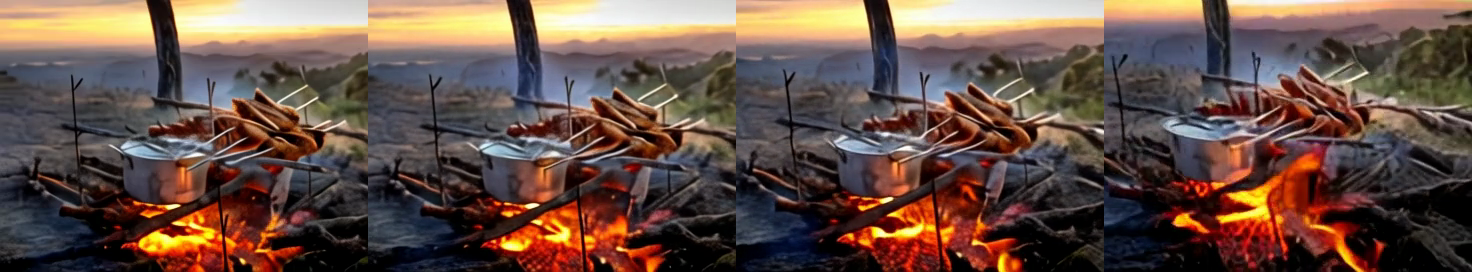} \\
 & \textit{``a bunch of food is cooking on a grill over an open fire''} \\
 \\
 \rotatebox[origin=l]{90}{~~~~~Wan} & \includegraphics[width=0.95\linewidth]{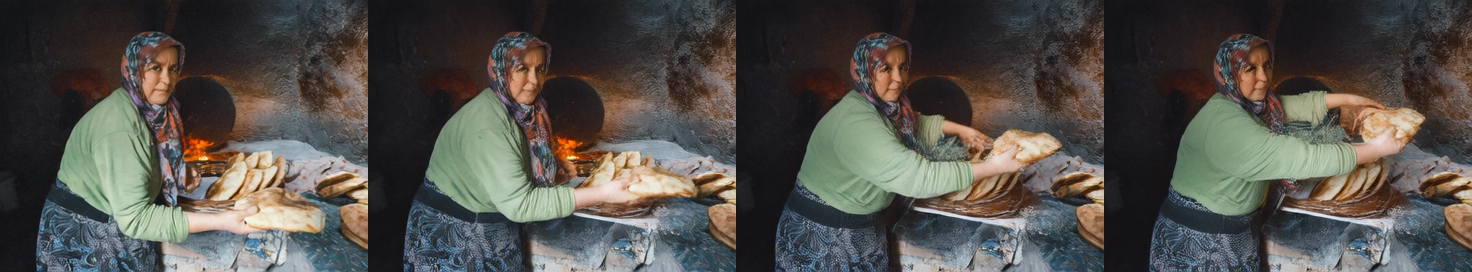} \\
 \rotatebox[origin=l]{90}{~~~+V-JEPA} & \includegraphics[width=0.95\linewidth]{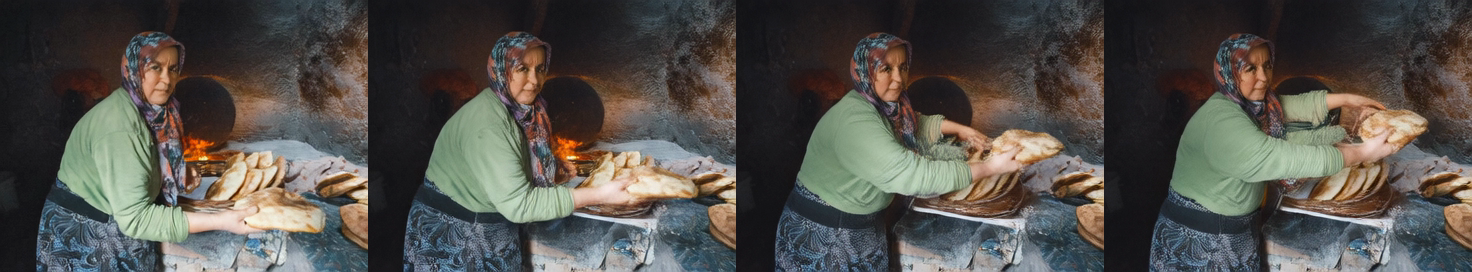} \\
 \rotatebox[origin=l]{90}{~~~+VCD} & \includegraphics[width=0.95\linewidth]{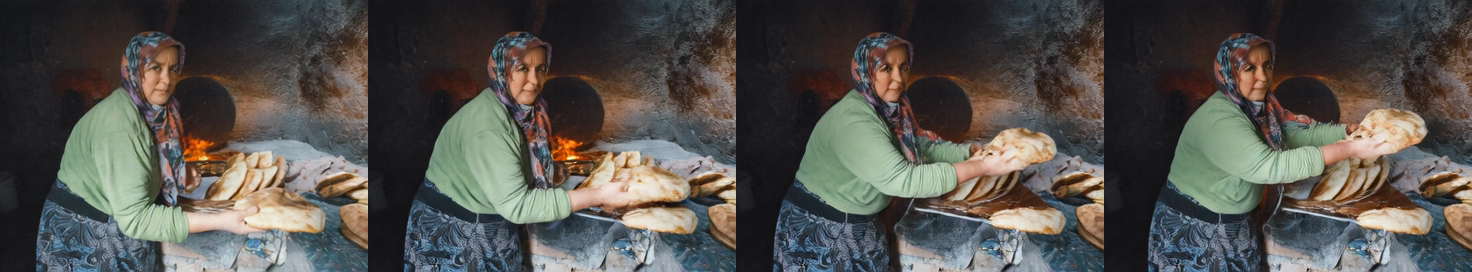} \\
 & \textit{``a woman is making bread in an oven''} 
\end{tabular}
\caption{
Results of video generation with VBench-I2V.
}
\label{fig:appendix-results-opensora-vbench_i2v}
\end{figure}

\begin{figure}[t]
\centering
\footnotesize
\tabcolsep=.4mm
\begin{tabular}{cc}
  \rotatebox[origin=l]{90}{~~Open-Sora} & \includegraphics[width=0.95\linewidth]{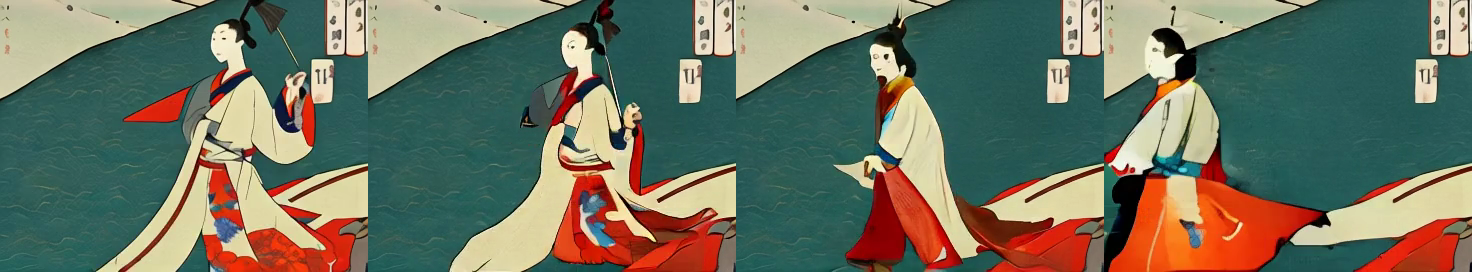} \\
 \rotatebox[origin=l]{90}{~~~+V-JEPA} & \includegraphics[width=0.95\linewidth]{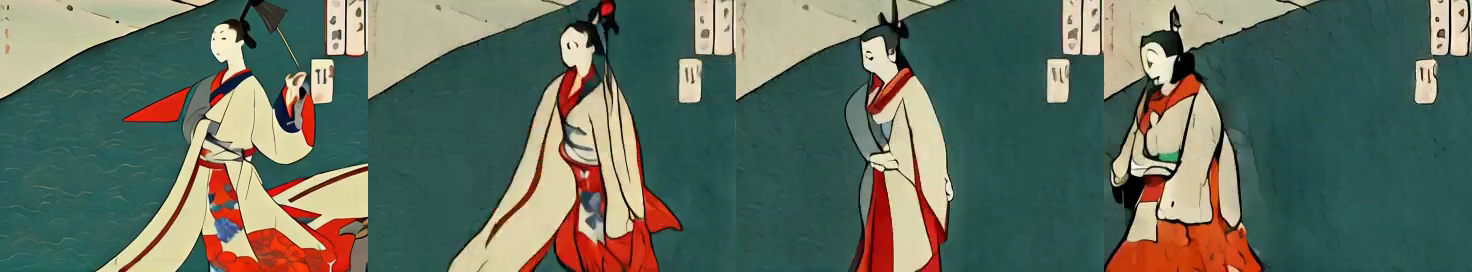} \\
 \rotatebox[origin=l]{90}{~~~+VCD} & \includegraphics[width=0.95\linewidth]{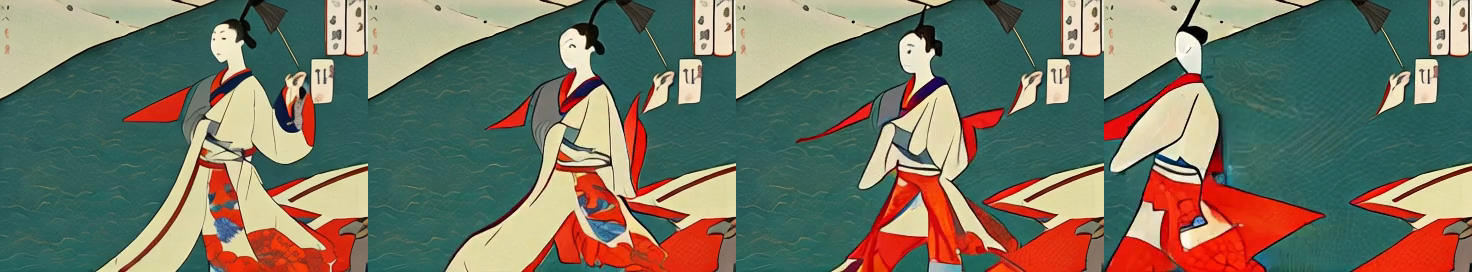} \\
  & \textit{``a person is walking forward, with the robe flowing behind''} \\
  \\
  \rotatebox[origin=l]{90}{~~~~~Wan} & \includegraphics[width=0.95\linewidth]{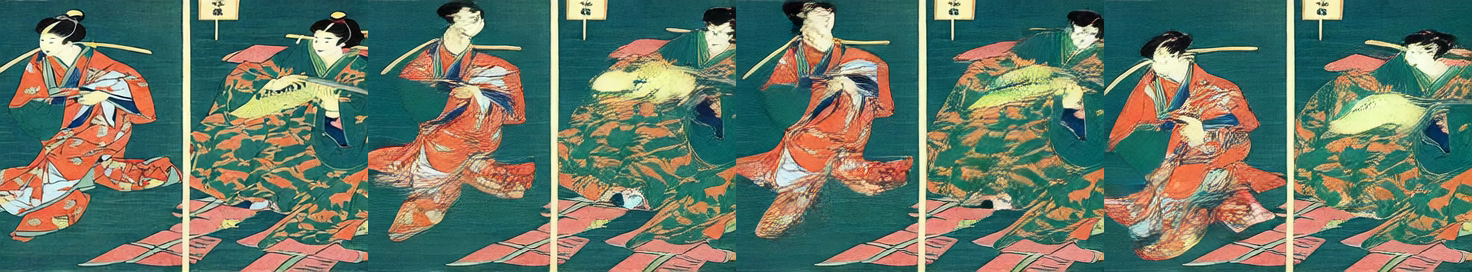} \\
 \rotatebox[origin=l]{90}{~~~+V-JEPA} & \includegraphics[width=0.95\linewidth]{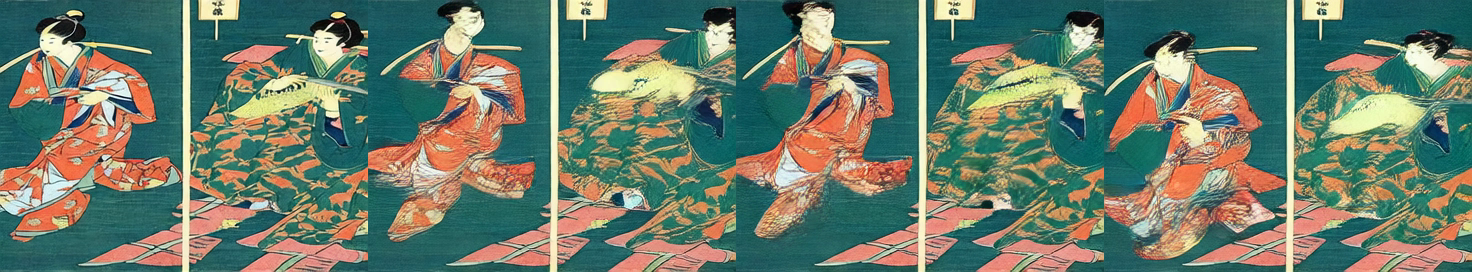} \\
 \rotatebox[origin=l]{90}{~~~+VCD} & \includegraphics[width=0.95\linewidth]{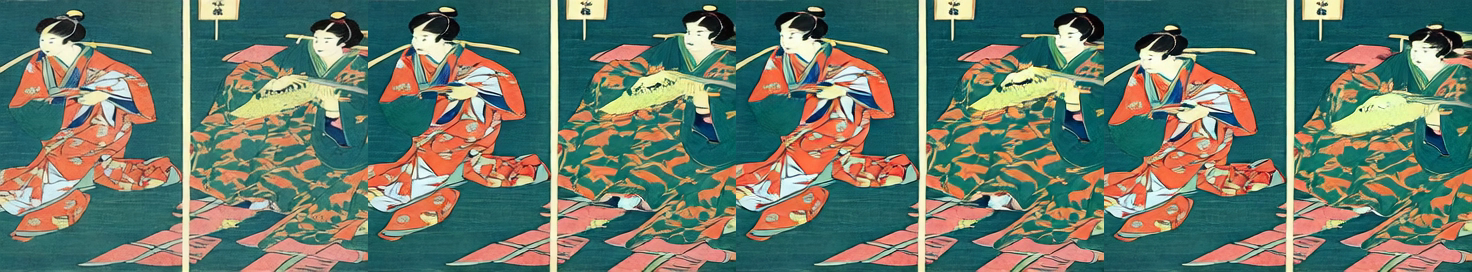} \\
  & \textit{``two people are dancing''}
\end{tabular}
\caption{
Results of video generation with AI-ArtBench.
}
\label{fig:appendix-results-opensora-ai_artbench}
\end{figure}

\begin{table*}[t]
  \centering  
  \scriptsize
  \tabcolsep=0.9mm
  \caption{VBench-I2V and Video-Text Alignment of Open-Sora and its fine-tuned models [\%].
  The means $\pm$ 95\% confidence intervals of five runs.
  A higher score indicates relatively better performance.
  The best and second best results are emphasized by \textbf{bold} and \underline{underlined} fonts, respectively.}
  \label{tab:quantitative-vbench_i2v}
  \begin{tabular}{lcccccccccc|c}
    \toprule
     & \textbf{I2V} & \textbf{I2V} & \textbf{Subject} & \textbf{Background} & \textbf{Temporal} & \textbf{Camera} & \textbf{Motion} & \textbf{Dynamic} & \textbf{Aesthetic} & \textbf{Imaging} & \textbf{Video-Text} \\
     & \textbf{Subject} & \textbf{Background} & \textbf{Consistency} & \textbf{Consistency} & \textbf{Flickering} & \textbf{Motion} & \textbf{Smoothness} & \textbf{Degree} & \textbf{Quality} & \textbf{Quality} & \textbf{Alignment} \\
    \midrule
    Open-Sora & \underline{94.25}\tiny{$\pm$0.21} & \underline{96.08}\tiny{$\pm$0.18} & \underline{96.27}\tiny{$\pm$0.21} & \underline{98.67}\tiny{$\pm$0.11} & \underline{98.98}\tiny{$\pm$0.05} & \underline{33.79}\tiny{$\pm$1.50} & \underline{99.05}\tiny{$\pm$0.05} & \underline{26.59}\tiny{$\pm$2.53} & \underline{54.42}\tiny{$\pm$0.35} & 62.75\tiny{$\pm$0.42} & \textbf{23.68}\tiny{$\pm$0.11} \\
    \midrule
    +V-JEPA & 92.81\tiny{$\pm$0.30} & 94.93\tiny{$\pm$0.29} & 93.83\tiny{$\pm$0.29} & 97.38\tiny{$\pm$0.16} & 98.48\tiny{$\pm$0.11} & \textbf{35.36}\tiny{$\pm$0.15} & 98.30\tiny{$\pm$0.13} & \textbf{29.92}\tiny{$\pm$2.56} & 51.67\tiny{$\pm$0.33} & \textbf{63.09}\tiny{$\pm$0.45} & 23.45\tiny{$\pm$0.11} \\
    \midrule
    +VCD (Ours) & \textbf{95.22}\tiny{$\pm$0.16} & \textbf{96.32}\tiny{$\pm$0.14} & \textbf{97.94}\tiny{$\pm$0.14} & \textbf{99.08}\tiny{$\pm$0.07} & \textbf{99.21}\tiny{$\pm$0.04} & 24.64\tiny{$\pm$1.37} & \textbf{99.27}\tiny{$\pm$0.03} & 17.56\tiny{$\pm$2.13} & \textbf{54.84}\tiny{$\pm$0.35} & \underline{63.04}\tiny{$\pm$0.41} & \underline{23.50}\tiny{$\pm$0.11} \\
    \bottomrule
    \toprule
    Wan & \underline{97.94}\tiny{$\pm$0.09} & \underline{98.76}\tiny{$\pm$0.04} & \underline{96.40}\tiny{$\pm$0.18} & \underline{98.87}\tiny{$\pm$0.06} & \underline{97.90}\tiny{$\pm$0.07} & \textbf{22.28}\tiny{$\pm$1.32} & \underline{98.61}\tiny{$\pm$0.05} & \underline{43.82}\tiny{$\pm$2.78} & 61.38\tiny{$\pm$0.35} & \underline{69.89}\tiny{$\pm$0.33} & \underline{23.39}\tiny{$\pm$0.11} \\
    \midrule
    +V-JEPA & 97.93\tiny{$\pm$0.09} & 98.75\tiny{$\pm$0.05} & 96.39\tiny{$\pm$0.18} & 98.85\tiny{$\pm$}0.06 & 97.88\tiny{$\pm$0.07} & \underline{22.18}\tiny{$\pm$1.32} & 98.60\tiny{$\pm$0.05} & \textbf{44.47}\tiny{$\pm$2.78} & \underline{61.40}\tiny{$\pm$0.35} & \textbf{69.92}\tiny{$\pm$0.33} & 23.38\tiny{$\pm$0.11} \\
    \midrule
    +VCD (Ours) & \textbf{98.20}\tiny{$\pm$0.07} & \textbf{98.78}\tiny{$\pm$0.04} & \textbf{97.12}\tiny{$\pm$0.15} & \textbf{99.02}\tiny{$\pm$0.05} & \textbf{98.18}\tiny{$\pm$0.06} & 20.31\tiny{$\pm$1.28} & \textbf{98.84}\tiny{$\pm$0.04} & 39.27\tiny{$\pm$2.73} & \textbf{61.55}\tiny{$\pm$0.35} & 69.86\tiny{$\pm$0.32} & \textbf{23.42}\tiny{$\pm$0.11} \\
    \bottomrule
  \end{tabular}
\end{table*}

\begin{table*}[t]
  \centering  
  \scriptsize
  \tabcolsep=0.5mm
  \caption{VideoScore of baseline models and their fine-tuned models in I2V-Bench and AI-ArtBench.
  A higher score indicates relatively better performance.}
  \label{tab:quantitative-videoscore}
  \begin{tabular}{lccccc|ccccc}
    \toprule
     & \multicolumn{5}{c}{\textbf{I2V-Bench}} & \multicolumn{5}{c}{\textbf{AI-ArtBench}} \\
     \cmidrule(rl){2-6} \cmidrule(rl){7-11}
     & \textbf{Visual} & \textbf{Temporal} & \textbf{Dynamic} & \textbf{Text-to-Video} & \textbf{Factual} & \textbf{Visual} & \textbf{Temporal} & \textbf{Dynamic} & \textbf{Text-to-Video} & \textbf{Factual} \\
     & \textbf{Quality} & \textbf{Consistency} & \textbf{Degree} & \textbf{Alignment} & \textbf{Consistency} & \textbf{Quality} & \textbf{Consistency} & \textbf{Degree} & \textbf{Alignment} & \textbf{Consistency} \\
    \midrule
    Open-Sora & 2.2287\tiny{$\pm$0.0065} & 2.2471\tiny{$\pm$0.0058} & 2.9191\tiny{$\pm$0.0021} & \underline{2.5222}\tiny{$\pm$0.0028} & 2.2198\tiny{$\pm$0.0067} & 1.9946\tiny{$\pm$0.0500} & 1.8848\tiny{$\pm$0.3863} & 2.7358\tiny{$\pm$0.0076} & \underline{2.3554}\tiny{$\pm$0.0032} & 1.7427\tiny{$\pm$0.0514} \\
    \midrule
    +V-JEPA & \underline{2.2603}\tiny{$\pm$0.0484} & \underline{2.3267}\tiny{$\pm$0.0036} & \textbf{2.9430}\tiny{$\pm$0.0262} & 2.4728\tiny{$\pm$0.0609} & \underline{2.2305}\tiny{$\pm$0.0438} & \underline{2.0138}\tiny{$\pm$0.0402} & \underline{1.9198}\tiny{$\pm$0.0333} & \textbf{3.1008}\tiny{$\pm$0.0308} & 2.3178\tiny{$\pm$0.0236} & \underline{1.8374}\tiny{$\pm$0.0461} \\
    \midrule
    +VCD (Ours) & \textbf{2.2907}\tiny{$\pm$0.0033} & \textbf{2.3427}\tiny{$\pm$0.0036} & \underline{2.9247}\tiny{$\pm$0.0032} & \textbf{2.5588}\tiny{$\pm$0.0030} & \textbf{2.2932}\tiny{$\pm$0.0035} & \textbf{2.1132}\tiny{$\pm$0.0456} & \textbf{1.9759}\tiny{$\pm$0.0303} & \underline{2.7778}\tiny{$\pm$0.0076} & \textbf{2.3970}\tiny{$\pm$0.0140} & \textbf{1.9131}\tiny{$\pm$0.0408} \\
    \bottomrule
    \toprule
    Wan & 2.2393\tiny{$\pm$0.0080} & 2.2588\tiny{$\pm$0.0079} & \underline{2.8796}\tiny{$\pm$0.0041} & 2.5120\tiny{$\pm$0.0061} & 2.2158\tiny{$\pm$0.0090} & 1.7749\tiny{$\pm$0.0362} & 1.5877\tiny{$\pm$0.0241} & \underline{2.8019}\tiny{$\pm$0.0134} & 2.3466\tiny{$\pm$0.0234} & \underline{1.4080}\tiny{$\pm$0.0249} \\
    \midrule
    +V-JEPA & \underline{2.2407}\tiny{$\pm$0.0080} & \underline{2.2591}\tiny{$\pm$0.0079} & \textbf{2.8799}\tiny{$\pm$0.0041} & \underline{2.5124}\tiny{$\pm$0.0061} & \underline{2.2163}\tiny{$\pm$0.0090} & \underline{1.7770}\tiny{$\pm$0.0365} & \underline{1.5912}\tiny{$\pm$0.0243} & \textbf{2.8042}\tiny{$\pm$0.0132} & \underline{2.3712}\tiny{$\pm$0.0245} & 1.4029\tiny{$\pm$0.0250} \\
    \midrule
    +VCD (Ours) & \textbf{2.2538}\tiny{$\pm$0.0084} & \textbf{2.2760}\tiny{$\pm$0.0081} & 2.8732\tiny{$\pm$0.0044} & \textbf{2.5297}\tiny{$\pm$0.0061} & \textbf{2.2257}\tiny{$\pm$0.0092} & \textbf{1.9159}\tiny{$\pm$0.0381} & \textbf{1.6718}\tiny{$\pm$0.0260} & 2.7422\tiny{$\pm$0.0147} & \textbf{2.4354}\tiny{$\pm$0.0217} & \textbf{1.5019}\tiny{$\pm$0.0274} \\
    \bottomrule
  \end{tabular}
\end{table*}

\begin{table*}[t]
  \centering  
  \footnotesize
  \tabcolsep=1.5mm
  \caption{Human evaluation results [\%].
  "Tie" indicates that annotators evaluated two videos are comparable.
  The results that showed statistically significant improvements with $p<0.001$ and $0.001\leq p < 0.005$ of the t-test are emphasized by \textbf{bold} and \underline{underlined} fonts, respectively.}
  \label{tab:quantitative-human}
  \begin{tabular}{c|ccccccccc}
    \toprule
     & \multicolumn{3}{c}{\textbf{I2V-Bench}} & \multicolumn{3}{c}{\textbf{VBench-I2V}} &  \multicolumn{3}{c}{\textbf{AI-ArtBench}} \\
     \cmidrule(rl){2-4} \cmidrule(rl){5-7} \cmidrule(rl){8-10}
                       & \textbf{Video-Text} & \textbf{Temporal} & \textbf{Motion} &  \textbf{Video-Text} & \textbf{Temporal} & \textbf{Motion} &  \textbf{Video-Text} & \textbf{Temporal} & \textbf{Motion} \\
     & \textbf{Alignment} & \textbf{Consistency} & \textbf{Naturalness}  & \textbf{Alignment} & \textbf{Consistency} & \textbf{Naturalness} & \textbf{Alignment} & \textbf{Consistency} & \textbf{Naturalness} \\
     \midrule
    \textbf{Open-Sora} & 17.78 & 16.89 & 12.00 & 12.67 & 18.22 & 7.11  & 10.44 & \textbf{28.00} & \textbf{15.11} \\
    \textbf{Tie}       & 66.67 & 62.44 & 74.89 & 72.00 & 60.44 & 83.78 & 81.33 & 69.11          & 82.89 \\
    \textbf{+V-JEPA}   & 15.56 & 20.67 & 13.11 & 15.33 & 21.33 & 9.11  & 8.22  & 2.89           & 2.00 \\
    \midrule
    \textbf{Open-Sora} & 12.22          & 9.11           & 16.67 & 10.44 & 6.00           &  4.22            & 5.33              & 2.89           & 2.67 \\
    \textbf{Tie}       & 66.44          & 58.00          & 64.89 & 77.78 & 71.78          & 86.44            & 83.78             & 70.89          & 84.89 \\
    \textbf{+VCD}      & \textbf{21.33} & \textbf{32.89} & 18.44 & 11.78 & \textbf{22.22} & \underline{9.33} & \underline{10.89} & \textbf{26.22} & \textbf{12.44} \\
    \midrule
    \textbf{V-JEPA}    & 18.89 & 5.33           & 12.00 & 11.78 & 4.44           & 4.89           & 15.78 & 7.78           & 3.56  \\
    \textbf{Tie}       & 60.44 & 58.00          & 72.89 & 75.56 & 60.67          & 83.78          & 65.56 & 50.00          & 75.33 \\
    \textbf{+VCD}      & 20.67 & \textbf{36.67} & 15.11 & 12.67 & \textbf{34.89} & \textbf{11.33} & 18.67 & \textbf{42.22} & \textbf{21.11} \\
    \bottomrule
    \toprule
    \textbf{Wan}       & 2.44  & 3.56  & 0.89  & 0.67  & 3.11  & 0.89  & 2.22  & 4.22  & 0.22 \\
    \textbf{Tie}       & 95.11 & 90.89 & 96.67 & 97.56 & 93.78 & 98.67 & 96.67 & 9.289 & 99.11 \\
    \textbf{+V-JEPA}   & 2.44  & 5.56  & 2.44  & 1.78  & 3.11  & 0.44  & 1.11 & 2.89 & 0.67 \\
    \midrule
    \textbf{Wan}       & 1.11           & 2.89           & 1.78           & 5.33              & 6.22           & 5.33  & 12.22 & 3.56           & 6.22 \\
    \textbf{Tie}       & 87.56          & 77.11          & 87.33          & 83.33             & 74.89          & 83.11 & 72.44 & 64.44          & 78.67 \\
    \textbf{+VCD}      & \textbf{11.33} & \textbf{20.00} & \textbf{10.89} & \underline{11.33} & \textbf{18.89} & 11.56 & 15.33 & \textbf{32.00} & \textbf{15.11} \\
    \midrule
    \textbf{V-JEPA}    & 2.67  & 3.11           & 1.78          & 4.89  & 2.22           & 3.11          & 13.11 & 4.89           & 4.89  \\
    \textbf{Tie}       & 91.78 & 82.67          & 90.67         & 87.33 & 80.67          & 87.78         & 70.22 & 60.89          & 77.78 \\
    \textbf{+VCD}      & 5.56  & \textbf{14.22} & \textbf{7.56} &  7.78 & \textbf{17.11} & \textbf{9.11} & 16.77 & \textbf{34.22} & \textbf{17.33} \\
    \bottomrule
  \end{tabular}
\end{table*}

\begin{table*}[t]
  \centering  
  \tabcolsep=1mm
  \caption{VideoScore of Open-Sora and +VCD w/ and w/o temporal weight $\frac{N-i+1}{N}$  in VBench-I2V dataset.
  A higher score indicates relatively better performance.
  The best and second best results are emphasized by \textbf{bold} and \underline{underlined} fonts, respectively.}
  \label{tab:appendix-quantitative-videoscore}
  \begin{tabular}{lccccc}
    \toprule
     & \textbf{Visual} & \textbf{Temporal} & \textbf{Dynamic} & \textbf{Video-Text} & \textbf{Factual} \\
     & \textbf{Quality} & \textbf{Consistency} & \textbf{Degree} & \textbf{Alignment} & \textbf{Consistency} \\
    \midrule
    Open-Sora & \underline{2.3517} & \underline{2.5481} & \underline{2.7279} & \underline{2.7384} & \underline{2.4220} \\
    \midrule
    +VCD w/o temporal weight & 2.2188 & 2.4470 & \textbf{2.7797} & 2.6977 & 2.2806 \\
    \midrule
    +VCD w/ temporal weight & \textbf{2.3865} & \textbf{2.5870} & 2.6935 & \textbf{2.7545} & \textbf{2.4535} \\
    \bottomrule
  \end{tabular}
\end{table*}

\section{Computational Efficiency of VCD}
We highlight the computational efficiency of our proposed Video Consistency Distance (VCD).
Specifically, VCD uses the shallow layers of VGG19~\cite{simonyan2015deepconvolutionalnetworkslargescale} with about 20 million parameters to extract frame features.
In contrast, the comparative method, V-JEPA~\cite{bardes2024revisiting}, employs a significantly larger network with about 1.3 billion parameters, approximately 65 times larger than that of VCD.
Thanks to the efficient design of VCD, it showed better results, as discussed in section~\ref{subsec:results}, than V-JEPA with fewer parameters.

\section{Additional Experiments and Results}
\subsection{Details of Experimental Settings}
\label{sec:appendix-details-settings}
\paragraph{Details of Evaluation Metrics}
For evaluation, we employed two evaluation benchmarks, VBench-I2V and VideoScore.
VBench-I2V comprises 10 evaluation dimensions, namely I2V Subject, I2V Background, Camera Motion, Subject Consistency, Background Consistency, Motion Smoothness, Dynamic Degree, Aesthetic Quality, Imaging Quality, and Temporal Flickering.
I2V Subject evaluates the consistency between the subject in the conditioning image and the corresponding subject in the generated video by calculating DINOv1~\cite{9709990} feature similarities.
I2V Background evaluates the consistency of the scene background between the conditioning image and the generated video frames by calculating DreamSim~\cite{fu2023dreamsim} feature similarities.
Camera Motion evaluates whether the generated video follows the camera motion described in the text prompt using CoTracker~\cite{karaev2023cotracker}.
Subject Consistency evaluates temporal consistency of the subject in a generated video throughout the whole video by calculating DINOv1~\cite{9709990} feature similarities.
Background Consistency evaluates temporal consistency of the background in the generated video throughout the whole video by calculating DreamSim~\cite{fu2023dreamsim} feature similarities.
Motion Smoothness evaluates whether the motion in the generated video is smooth using a video interpolation model~\cite{licvpr23amt}.
Dynamic Degree measures the proportion of videos that contain large motions using RAFT~\cite{teed2020raftrecurrentallpairsfield}.
Aesthetic Quality evaluates how the generated frames are artistic and beautiful using LAION Aesthetic predictor~\cite{aesthetic}.
Imaging Quality evaluates the distortion in the generated frames using MUSIQ~\cite{ke2021musiqmultiscaleimagequality}.
Temporal Flickering evaluates temporal consistency in local and high-frequency details by calculating the mean absolute difference between frames.
See more details in \cite{huang2024vbench++}.
Notably, VBench-I2V provides a cropping utility to match the input resolution requirements of video diffusion models.
As we generated a video with an approximate 4:3 aspect ratio, we cropped each conditioning image accordingly.

VideoScore evaluates videos with Visual Quality, Temporal Consistency, Dynamic Degree, Text-to-Video Alignment, and Factual Consistency, using the fine-tuned MantisIdefics2-8B~\cite{Jiang2024MANTISIM} with a human-annotated generated videos dataset of the above five metrics.
For each aspect, the dataset was annotated according to the following definitions, with a score range of 1 to 4.
Visual Quality evaluates the clarity, resolution, brightness, and color fidelity of the generated video.
Temporal consistency evaluates the consistency of objects or humans in the generated video over time.
Dynamic Degree evaluates the degree of dynamic changes in the generated video.
Text-to-Video Alignment evaluates how well the generated video content aligns with the input text prompt.
Factual Consistency evaluates whether the video content aligns with real-world facts and common-sense knowledge.
The definitions of each metric are also presented in Table 2 in \cite{he2024videoscore}.

\paragraph{Details of Human Evaluation Settings}
Figure~\ref{fig:appendix-ui} shows a screenshot of our user interface for human evaluation.
A conditioning image (left) and two generated video (middle and right) were presented along with a prompt.
Evaluators were tasked to choose whether video A or B is preferred or neutral.
When two videos were displayed side by side, their left/right order was randomized across trials to control for potential side bias.
Evaluators judged videos in terms of Temporal Consistency, Video-Text Alignment, and Motion Naturalness.
For Temporal Consistency, evaluators were asked which video remained faithful to the conditioning image across frames.
For Video-Text Alignment, they were asked which video was better aligned relative to the text prompt.
For Motion Naturalness, they were asked which video showed more natural movements.

Evaluators reached 82.9 percent agreement, indicating that the majority choice was consistent across evaluators, and 0.224 Fleiss' $\kappa$, which falls into the fair agreement range.
This discrepancy is expected since Fleiss' $\kappa$ corrects for chance agreement and is sensitive to skewed label distributions (in this case, frequent ''Tie`` votes).
Nevertheless, the high percent agreement suggests that the human evaluation results are reliable.

\subsection{Additional Results}
\label{sec:appendix-additional-results}
We provided additional qualitative results in Fig.~\ref{fig:appendix-results-opensora-i2v_bench}, \ref{fig:appendix-results-opensora-vbench_i2v}, and \ref{fig:appendix-results-opensora-ai_artbench}.
In the top part of Fig.~\ref{fig:appendix-results-opensora-i2v_bench}, Open-Sora unnaturally changed the color of the eyes and +V-JEPA changes the bangs.
In the bottom part of Fig.~\ref{fig:appendix-results-opensora-i2v_bench}, Wan and +V-JEPA generated distorted dogs.
The top part of Fig.~\ref{fig:appendix-results-opensora-vbench_i2v} showed that Open-Sora generated significantly different frames from the conditioning image and +V-JEPA generated distorted frames.
The bottom part of Fig.~\ref{fig:appendix-results-opensora-vbench_i2v} showed that the region between the woman's arms distorted in the generated videos by Wan and +V-JEPA.
In the top part of Fig.~\ref{fig:appendix-results-opensora-ai_artbench}, Open-Sora and +V-JEPA significantly distorted the clothing.
In the bottom part of Fig.~\ref{fig:appendix-results-opensora-ai_artbench}, Wan and +V-JEPA distorted two people.
In contrast to these results, +VCD generated natural videos following the text prompt compared to Open-Sora and +V-JEPA.

We summarized VBench-I2V, VideoScore, and the human evaluation results in Table~\ref{tab:quantitative-vbench_i2v}, \ref{tab:quantitative-videoscore}, and \ref{tab:quantitative-human}.
The results are identical to Fig.~\ref{fig:results-human-evaluation}.

\subsection{Ablation Study}
\label{sec:appendix-ablation}
\paragraph{Temporal Weight}
As described in Section~\ref{subsec:vcd}, we introduced a temporal weight $\frac{N-i+1}{N}$ for VCD to prevent generating a still image.
We evaluated its effectiveness.

Table~\ref{tab:appendix-quantitative-videoscore} shows the results of VideoScore for the following three models: (1) Open-Sora (2) fine-tuned Open-Sora using VCD without a temporal weight (3) fine-tuned Open-Sora using VCD with a temporal weight.
+VCD w/o a temporal weight showed worse results in Visual Quality, Temporal Consistency, Video-Text Alignment, and Factual Consistency than Open-Sora and +VCD w/ a temporal weight.
These results indicate that a temporal weight restricts degrading generated video qualities.

\paragraph{Wasserstein Distance and Frequency Space}
We design VCD to calculate the Wasserstein Distance between a conditioning image and a generated frame in frequency space.
To evaluate the effectiveness of the design, we fine-tuned Open-Sora with L2 loss (instead of Wasserstein distance) in frequency space and with Wasserstein distance loss in feature space (instead of frequency space) as follows: 

\begin{equation}
\begin{split}
\mathrm{VCD}_{L2} = \frac{N-i+1}{N}(||\mathcal{A}_{\mathrm{E}(x_{\mathrm{cnd}})},\mathcal{A}_{\mathrm{E}(x_{i})}|| \\
+ ||\mathcal{A}_{\mathrm{E}(x_{\mathrm{cnd}}},\mathcal{A}_{\mathrm{E}(x_{i})}||),
\end{split}
\end{equation}

\begin{equation}
\begin{split}
\mathrm{VCD}_{\mathrm{Feat.}} = \frac{N-i+1}{N}(\mathrm{WD}(\mathrm{E}(x_{\mathrm{cnd}}),\mathrm{E}(x_{i})) \\
+ \mathrm{WD}(\mathrm{E}(x_{\mathrm{cnd}},\mathrm{E}(x_{i})))).
\end{split}
\end{equation}

\begin{figure}[t]
\centering
\footnotesize
\tabcolsep=.4mm
\begin{tabular}{cc}
 \rotatebox[origin=l]{90}{~~~~~+L2} & \includegraphics[width=0.95\linewidth]{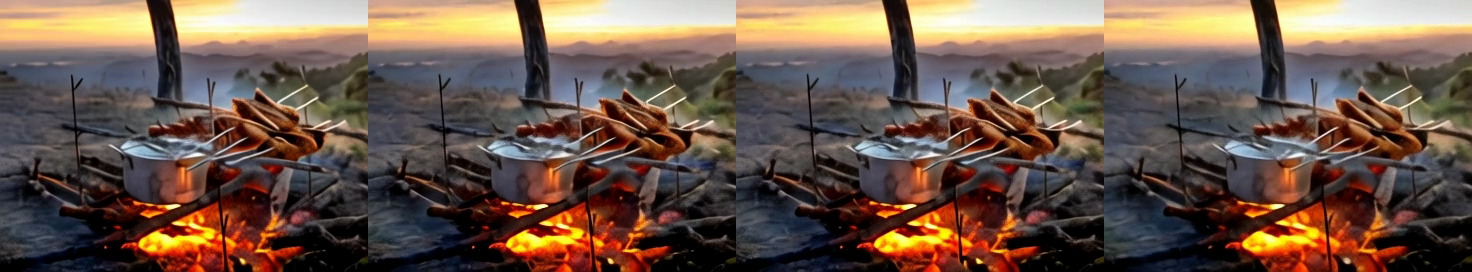} \\
 \rotatebox[origin=l]{90}{~~~+Feat.} & \includegraphics[width=0.95\linewidth]{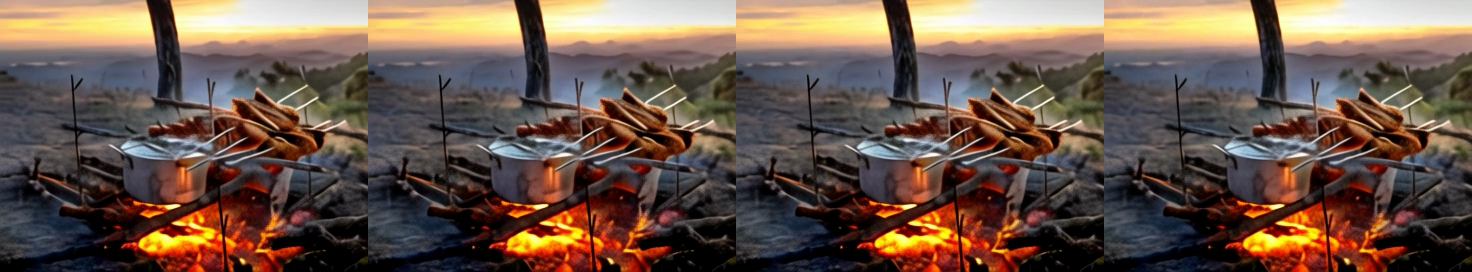} \\
 \rotatebox[origin=l]{90}{~~~+VCD} & \includegraphics[width=0.95\linewidth]{./figures/a_bunch_of_food_is_cooking_on_a_grill_over_an_open_fire-opensora_400steps_vcd.png} \\
 & \textit{``a bunch of food is cooking on a grill over an open fire''} 
\end{tabular}
\caption{
Results of video generation with VBench-I2V.
+L2: Generated frames by Open-Sora fine-tuned with L2 loss (instead of Wasserstein distance) in frequency space.
+Feat.: Generated frames by Open-Sora fine-tuned with Wasserstein distance loss in feature space.
}
\label{fig:appendix-ablation-study}
\end{figure}

Also, we evaluated these models with VBench-I2V and observed significantly lower Dynamic Degree scores than +VCD (i.e., 17.56\% in Table~\ref{tab:quantitative-vbench_i2v}), 0.00\%, and 2.11\%, respectively.
These results support our design choice in VCD, which helps prevent the model from generating still images.
We provide examples where +VCD showed flickering flames, while the others showed no motion in Fig.~\ref{fig:appendix-ablation-study}.

\end{document}